\documentclass{article} 
\usepackage{iclr2027_conference,times}

\usepackage{amsmath,amsfonts,bm}

\def\eqref#1{equation~\ref{#1}}

\def\1{\bm{1}}

\DeclareMathAlphabet{\mathsfit}{\encodingdefault}{\sfdefault}{m}{sl}
\SetMathAlphabet{\mathsfit}{bold}{\encodingdefault}{\sfdefault}{bx}{n}

\usepackage{hyperref}
\usepackage{url}

\usepackage{booktabs}
\usepackage{float}
\usepackage{graphicx} 
\usepackage{multirow}
\usepackage{colortbl} 
\usepackage[dvipsnames]{xcolor}
\usepackage{tikz}
\usepackage{pgfplots}
\pgfplotsset{compat=1.18}
\usepackage{amsmath} 
\hypersetup{hidelinks}
\usepackage{subcaption}
\newcommand{\lzx}[1]{\textcolor{blue}{#1}}
\usepackage{mdframed}  
\newmdenv[
  backgroundcolor=gray!12,
  linewidth=0pt,
  innerleftmargin=10pt,
  innerrightmargin=10pt,
  innertopmargin=8pt,
  innerbottommargin=8pt,
  skipabove=6pt,
  skipbelow=6pt,
]{promptbox}

\usetikzlibrary{calc}

\definecolor{gridline}{HTML}{D8D8D8}
\definecolor{axiscolor}{HTML}{555555}
\definecolor{labelgray}{HTML}{333333}
\definecolor{subgray}{HTML}{000000}
\definecolor{bubbleA}{HTML}{81C39E}
\definecolor{bubbleB}{HTML}{70A1D5}
\definecolor{bubbleC}{HTML}{D78E73}
\definecolor{bubbleD}{HTML}{D78E73}

\usepackage{array}

\newcommand{\revisioncolor}{ForestGreen}

\title{ShadowPEFT: Shadow Network for Parameter Efficient Fine-Tuning}

\author{
    Xianming Li\textsuperscript{\rm 1},
    Zongxi Li\textsuperscript{\rm 2}\thanks{\ Equal contribution. \ $\dagger$ Corresponding Author.},
    Tsz-fung Andrew Lee\textsuperscript{\rm 2},
    Jing Li\textsuperscript{\rm 1}$^\dagger$
    , Haoran Xie\textsuperscript{\rm 2} , Qing Li\textsuperscript{\rm 1}  \\
    \textsuperscript{\rm 1} PolyU\ \ \ \ 
    \textsuperscript{\rm 2} School of Data Science,
    Lingnan University, Hong Kong SAR
    \\ \\
  \texttt{xianming.li@connect.polyu.hk}, \\
  \texttt{zongxili@ln.edu.hk}, \texttt{jing-amelia.li@polyu.edu.hk}\\
 }

\iclrfinalcopy 
\begin{document}

\maketitle

\begin{abstract}
Popular low-rank parameter-efficient fine-tuning (PEFT) methods represent adaptation as separate updates to selected backbone weights, without maintaining an explicit task-specific state that is updated and reused across depth. These updates also require the backbone at inference and therefore cannot operate as standalone predictors.
We propose ShadowPEFT, which consolidates trainable adaptation into a modular shadow component centered on a compact shadow model and lightweight Transformer layer-specific coupling modules. 
A persistent shadow state refines the frozen backbone representations and is updated from them in an interactive manner. 
Because the shadow model is trained as a complete predictor, it can be detached for shadow-only inference without executing the base model and can be initialized from a pretrained model. 
Experiments on text and image generation and understanding benchmarks show that ShadowPEFT matches or outperforms LoRA and DoRA under comparable trainable-parameter budgets. Additional analyses on shadow pretraining, cross-dataset transfer, parameter scaling, inference latency, and system-level evaluation suggest that centralized layer-space adaptation is a competitive and flexible alternative to conventional low-rank PEFT.
The code is available at: \url{https://github.com/ShadowLLM/shadow-peft}.

\end{abstract}

\section{Introduction}
\begin{figure}[htbp!]
 \centering
 \includegraphics[width=0.98\linewidth]{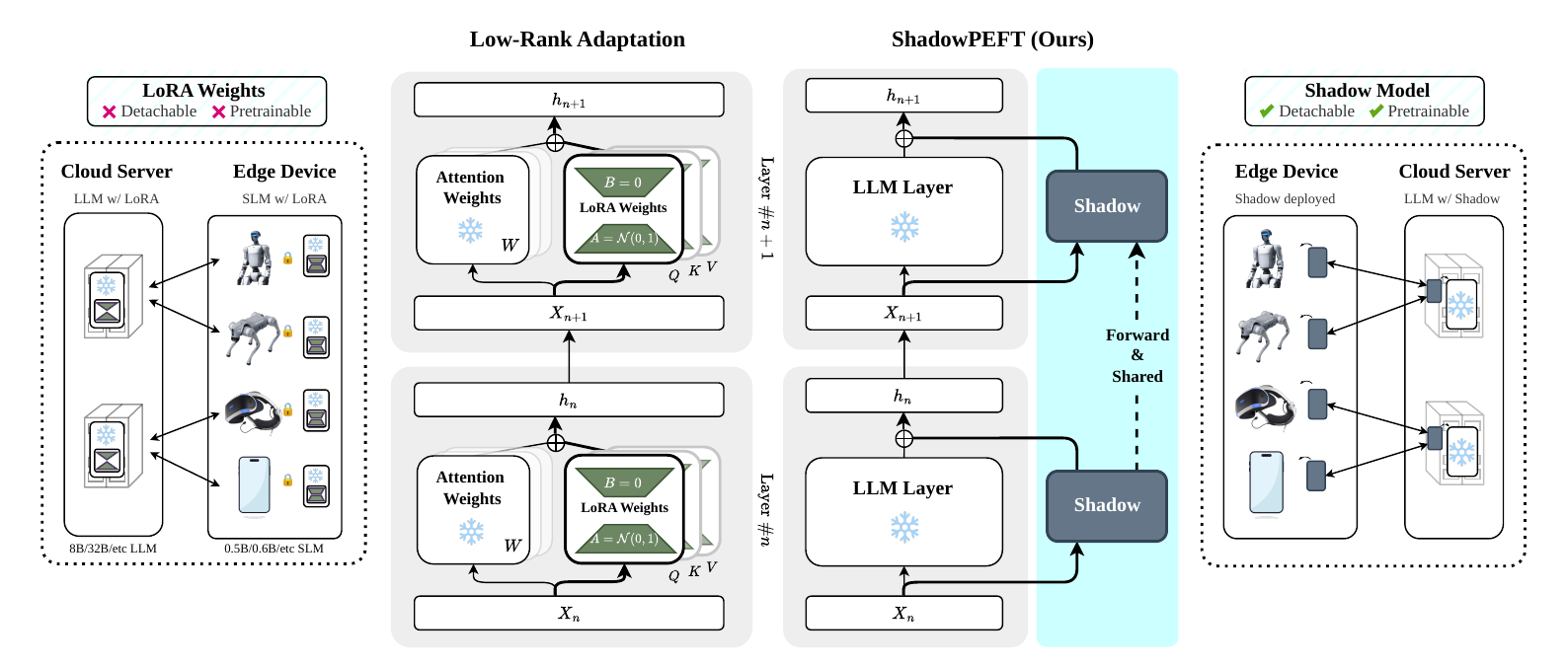}
 \caption{Comparison between (Left) conventional LoRA configuration and (Right) our proposed ShadowPEFT. LoRA-style methods adapt the backbone through linear-specific low-rank weight updates, whereas ShadowPEFT carries a persistent shadow state through depth. The shadow state can be attached to the backbone or detached for lightweight deployment.}
 \label{fig:teaser}
\end{figure}

Parameter-efficient fine-tuning (PEFT) trains a small set of task-specific parameters while keeping most backbone parameters frozen to make large-scale model adaptation computationally feasible \citep{Han2024ParameterEfficientFF}. Representative approaches include prompt and prefix tuning \citep{li2021prefix}, adapter modules \citep{houlsby2019parameter}, and low-rank methods such as LoRA \citep{hu2022lora}, QLoRA \citep{dettmers2023qlora}, and DoRA \citep{liu2024dora}. LoRA-style methods are especially popular for large language models because they are simple and broadly compatible. 
They encode adaptation, however, as separate local modifications to selected backbone weights. 
Although backbone hidden representations naturally propagate through depth, the adaptation mechanism itself does not maintain a task-specific state that is explicitly carried, updated, and reused across layers.
Cross-layer coordination of adaptation therefore emerges only indirectly through the backbone computation, while the trainable updates remain distributed across layers and dependent on the backbone at inference. 
This motivates a complementary question: \textbf{can task adaptation be accompanied by an explicit depth-wise state that coordinates layer representations?}

We propose \textbf{ShadowPEFT}, which consolidates trainable adaptation into a modular shadow component coupled to a frozen backbone. 
The component combines a compact shadow model (architecturally similar to the base model but scaled down) with lightweight, layer-specific injection and update modules. 
The shadow model initializes and maintains a task state; at each Transformer layer, this state refines the backbone representation, and the resulting backbone state updates the shadow state for the next layer. 
This interactive coupling mechanism gives later refinements access to both accumulated shadow and the latest backbone representation, allowing more effective task adaptation through progressively coordinated representation learning across depth. 

This design offers two complementary capabilities. 
First, as the shadow model is trained as a functional predictor, it can be detached and deployed without the base model. 
This feature supports deployment patterns such as edge–cloud collaboration, in which the compact shadow model handles simple requests locally and delegates more complex requests to an attached cloud model. Furthermore, since the Shadow module is a self-contained functional model, it can be initialized from another pretrained model. For example, a smaller model such as Qwen3~0.5B can serve as the Shadow model for a larger backbone such as Qwen3~8B, enabling the larger model to be adapted by training its smaller counterpart. In this configuration, the smaller model learns to refine the representations of the larger one, allowing adaptation capacity to be reused across model scales. It expands PEFT beyond lightweight parameter injection toward reusable, cross-scale adaptation dynamics.

We evaluate ShadowPEFT on six text and image generation and understanding benchmarks, including MMLU (text) \citep{hendryckstest2021}, GSM8K (text) \citep{cobbe2021gsm8k}, SQuADv2 (text) \citep{rajpurkar2018squadv2}, DreamBooth (image) \citep{ruiz2023dreambooth}, Amazon Review (text) \citep{keung2020amazonreview}, and 20 Newsgroups (text) \citep{twenty_newsgroups}. 
Across Qwen backbones at three scales, compact ShadowPEFT is competitive with LoRA and DoRA under closely matched trainable-parameter budgets, while enabling modular deployment modes that are not naturally supported by conventional low-rank PEFT.
Our contributions are as follows:

1. We introduce ShadowPEFT, a modular representation-space PEFT framework that organizes adaptation around a persistent shadow state and couples it to a frozen backbone through shadow-to-base injection and base-to-shadow gated updates.

2. We provide a detachable shadow model that supports both attached refinement and shadow-only inference, enabling deployment patterns such as edge--cloud collaboration while separating architectural availability from standalone predictive quality.

3. Across multiple benchmarks and backbone scales, ShadowPEFT achieves competitive performance relative to LoRA and DoRA under comparable trainable-parameter budgets, while demonstrating stronger out-of-distribution transfer in our cross-dataset evaluation. 

\section{Related Work}
\paragraph{Revisit parameter-efficient fine-tuning.}
PEFT adapts pretrained models while updating only a small fraction of their parameters. Prompt Tuning \citep{lester2021power}, Prefix Tuning \citep{li2021prefix}, and P-Tuning \citep{liu2022p} optimize continuous prompts while leaving the backbone unchanged. Adapter methods instead insert trainable bottleneck transformations into selected Transformer layers \citep{houlsby2019parameter}. Subsequent work improves adapter composition, transfer, or parameter sharing, including AdapterSoup \citep{chronopoulou2023adaptersoup}, Tiny-Attention Adapter \citep{zhao2022tiny}, and Compacter \citep{karimi2021compacter}. 
Standard adapters generally use layer-specific parameters, for instance, Compacter shares hypercomplex ``slow'' factors while retaining layer-specific rank-one factors \citep{karimi2021compacter}. 
ShadowPEFT does not tie its injection and update parameters across depth. Instead, successive layers are connected by an evolving shadow state, making state persistence---rather than parameter sharing---the source of depth-wise coordination.

\paragraph{Low-rank adaptation.}
LoRA \citep{hu2022lora} freezes pretrained weights and learns low-rank updates for selected linear projections, substantially reducing the number of trainable parameters. Extensions improve rank allocation (AdaLoRA \citep{zhangadaptive} and DyLoRA \citep{valipour2023dylora}), initialization (LoRA-GA \citep{wang2024lora}, PiSSA \citep{meng2024pissa}, and OLoRA \citep{buyukakyuz2024olora}), quantization (QLoRA \citep{dettmers2023qlora} and QA-LoRA \citep{xu2023qa}), and multi-task or compositional adaptation \citep{che2026lora,li2024mixlora,yang2025mtl,hsu2024safe}. 
These methods are effective, but their trainable adaptation remains distributed across layer-local weight updates. 
Although backbone activations  propagate across layers, the adaptation mechanism does not explicitly maintain a task-specific state that is updated and reused through depth. 
ShadowPEFT instead organizes adaptation around a persistent state operating on layer-level representations, while retaining lightweight layer-specific coupling transformations.

\paragraph{Side networks and two-way cross-pathway coupling.}
Side-Tuning trains a lightweight side network and additively fuses its output with that of a frozen base network \citep{zhang2020sidetuning}. Ladder Side-Tuning (LST) exposes intermediate frozen-backbone activations to a separate ladder network through shortcut connections, enabling training without back-propagating through the backbone \citep{sung2022lst}. These methods establish that a trainable side network can consume frozen-backbone features efficiently. 
ShadowPEFT is related in maintaining an auxiliary pathway, but its dataflow uses two-way coupling at every depth: shadow-to-base injection modifies the input to the next frozen layer, and base-to-shadow gated update refreshes the state used by later injections. 
Consequently, attached ShadowPEFT executes in dependency order rather than as two concurrently parallel networks, and unlike LST its training objective is not to eliminate back-propagation through the attached base computation. We view this coupled state transition, rather than the mere presence of a side network, as the relevant conceptual distinction.

\paragraph{Detachable adaptation pathways.} 
Low-rank weight updates and conventional adapters require the frozen backbone at inference because they do not form complete predictors on their own. 
ShadowPEFT instead equips the shadow backbone with a task prediction head and directly supervises its output during joint training. The same checkpoint therefore supports attached inference with the coupled base--shadow system and detached inference with only the shadow backbone and head. 
This functional detachability distinguishes ShadowPEFT from weight-update PEFT; its practical value must be assessed through shadow-only quality and deployment cost.

\section{Our ShadowPEFT Framework}
\label{sec:methodology}
We introduce \textbf{ShadowPEFT}, a PEFT framework that consolidates trainable task adaptation into a modular \emph{shadow component} coupled to a frozen pretrained LLM (the \emph{base model}). 
Unlike LoRA-style methods 
(\textcolor{blue}{Figure~\ref{fig:teaser}(left)}), ShadowPEFT organizes adaptation around a persistent \emph{shadow state} carried through depth (\textcolor{blue}{Figure~\ref{fig:teaser}(right)}). 

\subsection{ShadowPEFT Overview}
\label{sec:shadow_model}

Each layer of ShadowPEFT consists of three operations: shadow injection, frozen base encoding, and shadow update. Let the frozen base consist of $L$ Transformer decoder layers. Given an input sequence $\mathbf{x}$, let $\mathbf{h}_{out}^{(\ell)} \in \mathbb{R}^{ T \times d}$ denote the hidden state of the $\ell$-th LLM decoder layer (base layer), where $T$ is the sequence token length and $d$ is the hidden dimension. The \emph{shadow state}
$\mathbf{s}^{(\ell)} \in \mathbb{R}^{T \times d}$ is the persistent, task-adaptive state at depth $\ell$. The initial shadow state $\mathbf{s}^{(0)}$ is produced by a \emph{shadow backbone} $f_{\text{shadow}}$:
\begin{equation}
  \mathbf{s}^{(0)} = f_{\text{shadow}}\!\left(\mathbf{x};\,\theta_{\text{shadow}}\right).
\end{equation}
For each base layer $\ell \geq 1$, ShadowPEFT repeats three operations. First, shadow injection uses the current shadow state $\mathbf{s}^{(\ell-1)}$ to refine the previous base hidden state $\mathbf{h}_{out}^{(\ell-1)}$. Second, the frozen base layer $f^{(\ell)}_{\text{base}}$ encodes the refined representation and produces $\mathbf{h}_{out}^{(\ell)}$. Third, the shadow update module advances the shadow state from $\mathbf{s}^{(\ell-1)}$ to $\mathbf{s}^{(\ell)}$ using the newly produced base representation. 
These operations are dependency-ordered in a forward pass: injection $\rightarrow$ base encoding $\rightarrow$ update at each depth, rather than concurrently parallel branches. 
More details about our shadow-backbone designs are provided in \textcolor{blue}{Appendix \ref{apd:shadow_backbone_detail}}.

\begin{figure}
    \centering
    \includegraphics[width=\linewidth]{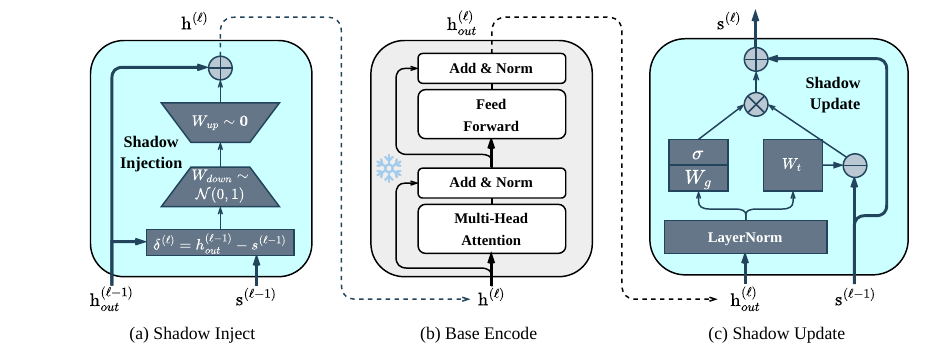}
    \caption{Architecture of ShadowPEFT. \textbf{(a) Shadow Injection Module.}
            The discrepancy $\boldsymbol{\delta}^{(\ell)}$
            is projected through a low-rank bottleneck
            ($W_{\!\mathrm{down}}{\sim}\mathcal{N}(0,\sigma^2)$,\;$W_{\!\mathrm{up}}{=}0$)
            and added back to the base hidden state. \textbf{(b) Base Encoding Module.} The frozen base layer encodes the refined representation.
            \textbf{(c) Shadow Update Module.}
            The base output $\mathbf{h}^{(\ell)}_{\mathrm{out}}$ is LayerNorm-normalised, then
            split into a transform $W_t$ and a sigmoid gate $\sigma(W_g)$;
            their product updates the shadow state via a  gated residual.
            }
    \label{fig:injection-update}
\end{figure}

\subsection{Shadow Injection Module}
\label{sec:injection}
The \textit{Shadow Injection Module} carries the shadow-to-base direction of the coupling. Its goal is to inject task-specific corrections into the hidden states of the backbone by comparing the current representation of the backbone with the evolving shadow state. Unlike LoRA-style methods, 
ShadowPEFT performs adaptation in the transformer decoder layer: the persistent shadow state provides the depth-dependent reference, and each layer-specific injection module learns how to use the base--shadow discrepancy to refine the next backbone-layer input.
At base layer $\ell$, the module takes the previous backbone output $\mathbf{h}_{out}^{(\ell-1)}$ and the current shadow state $\mathbf{s}^{(\ell-1)}$ as inputs, as illustrated in \textcolor{blue}{Figure \ref{fig:injection-update}(a)}. Since $\mathbf{s}^{(\ell-1)}$ has been initialized from the input and progressively updated through earlier layers, it encodes task-relevant information accumulated along the shadow pathway. 
The discrepancy between the two states therefore provides a compact signal indicating how the frozen backbone representation should be adjusted before entering layer $\ell$:
\begin{align}
   \boldsymbol{\delta}^{(\ell)}
     &= \mathbf{h}_{out}^{(\ell-1)} - \mathbf{s}^{(\ell-1)}. \label{eq:delta}
\end{align}
Rather than injecting this raw discrepancy to the backbone directly, ShadowPEFT passes it through a lightweight trainable bottleneck to obtain a refined task-specific correction:
$
   \tilde{\boldsymbol{\delta}}^{(\ell)}
     = \operatorname{Dropout}\!\left(\boldsymbol{\delta}^{(\ell)}
        \mathbf{W}^{(\ell)}_{\mathrm{down}}\right)
        \mathbf{W}^{(\ell)}_{\mathrm{up}}, \label{eq:proj}
$
where $\mathbf{W}^{(\ell)}_{\mathrm{down}} \in \mathbb{R}^{d \times r}$ and
$\mathbf{W}^{(\ell)}_{\mathrm{up}} \in \mathbb{R}^{r \times d}$ are layer-dependent trainable projection matrices with rank $r \ll d$.\footnote{\;Inspired by LoRA \citep{hu2022lora}, $\mathbf{W}^{(\ell)}_{\mathrm{down}}$ is initialized from $\mathcal{N}(0,\,\sigma^2)$ and $\mathbf{W}^{(\ell)}_{\mathrm{up}}$ is zero-initialized. This guarantees that $\tilde{\boldsymbol{\delta}}^{(\ell)} = \mathbf{0}$ at initialization, so the frozen base model remains unchanged at start of training.}
This bottleneck preserves parameter efficiency while preventing the model from directly copying the base--shadow difference into the backbone. Instead, it learns which components of the discrepancy are useful for adaptation, making the injected signal a learned refinement rather than a direct overwrite.
Finally, the correction is added to the backbone through a residual connection:
\begin{equation}
  \mathbf{h}^{(\ell)}
    = \mathbf{h}_{out}^{(\ell-1)}
       + \alpha\,\tilde{\boldsymbol{\delta}}^{(\ell)}, \label{eq:inject}
\end{equation}
where $\alpha>0$ controls the injection strength. The corrected representation $\mathbf{h}^{(\ell)}$ is then passed into base layer $\ell$ to produce $\mathbf{h}_{out}^{(\ell)}$, as illustrated in \textcolor{blue}{Figure~\ref{fig:injection-update}(b)}.

\subsection{Shadow Update Module}
\label{sec:update}
The \textit{Shadow Update Module} carries the base-to-shadow direction of the coupling and keeps the shadow state aligned with the frozen backbone as representations evolve across depth. Since the injection module relies on the discrepancy between the hidden state of the backbone and the shadow state, the shadow cannot remain fixed after initialization. Instead, it should absorb useful information from each backbone layer while preserving its accumulated task-specific context. This allows the shadow state to serve as a stable yet adaptive reference for subsequent injections.
To update the shadow state after base layer $\ell$, we use a gated residual mechanism, as illustrated in \textcolor{blue}{Figure~\ref{fig:injection-update}(c)}, that moves the previous shadow state $\mathbf{s}^{(\ell-1)}$ toward a candidate representation derived from the current backbone output $\mathbf{h}^{(\ell)}_{\mathrm{out}}$:
\begin{align}
  \mathbf{t}^{(\ell)}
    &= T^{(\ell)}\!\left(\mathbf{h}^{(\ell)}_{\mathrm{out}}\right), \label{eq:transform} \\
  \mathbf{g}^{(\ell)}
    &= \sigma\!\left(G^{(\ell)}\!\left(\mathbf{h}^{(\ell)}_{\mathrm{out}}\right)\right),
       \label{eq:gate} \\
  \mathbf{s}^{(\ell)}
    &= \left(1-\mathbf{g}^{(\ell)}\right)\odot \mathbf{s}^{(\ell-1)}+\mathbf{g}^{(\ell)}\odot \mathbf{t}^{(\ell)},
       \label{eq:update}
\end{align}
\noindent
where $\odot$ denotes element-wise multiplication and $\sigma$ is the sigmoid function. Here, $T^{(\ell)}(\cdot)$ maps the current backbone output into a candidate shadow representation, while $G^{(\ell)}(\cdot)$ predicts an element-wise gate that controls how much new information is incorporated into the shadow state. The implementations of $T^{(\ell)}(\cdot)$ and $G^{(\ell)}(\cdot)$ are provided in \textcolor{blue}{Appendix~\ref{apd:gated_residual_update}}.

This update can be viewed as an interpolation between the previous shadow state and the current layer-induced candidate state. The gate controls the balance between retaining information accumulated from earlier layers and incorporating information from the current backbone layer. Similar to GRU-style updates \citep{Cho2014GRU}, this mechanism helps prevent abrupt state replacement, stabilizes optimization, and reduces the risk of shadow collapse.


\subsection{Base-Shadow Joint Training}
\label{sec:joint_learning}
ShadowPEFT is trained end-to-end with a joint loss while keeping the base model frozen. The trainable components include the shadow backbone, the injection projections, the update networks, and the shadow prediction head. 

\paragraph{Causal language modeling objective.}
For causal language modeling, given shifted target labels $\mathbf{y}$,
the training loss is defined as:
\begin{equation}
  \mathcal{L} =
    \mathcal{L}_{\mathrm{CE}}\!\left(
      \mathbf{h}_{\mathrm{base}}\,\mathbf{W}_{\mathrm{lm}},\; \mathbf{y}
    \right)
    + \lambda\,
    \mathcal{L}_{\mathrm{CE}}\!\left(
      \mathbf{s}^{(0)}\,\mathbf{W}_{\mathrm{shadow}},\; \mathbf{y}
    \right),
  \label{eq:loss_lm}
\end{equation}
where $\mathbf{h}_{\mathrm{base}}$ is the final hidden state of the injected base model,
$\mathbf{W}_{\mathrm{lm}}$ and $\mathbf{W}_{\mathrm{shadow}}$ are the base and shadow
language-model heads, 
$\mathcal{L}_{\mathrm{CE}}$ is the standard next-token cross-entropy loss, and
$\lambda$ is a scalar hyperparameter (default $\lambda = 0.05$).

\paragraph{Sequence classification objective.}
For sequence classification with $C$ classes, following \cite{li2026ls}, the shadow hidden state $\mathbf{s}^{(L)} \in \mathbb{R}^{B \times T \times d}$ is reduced to a pooled vector. 
This pooled representation is then scored by a shadow classifier head
$\mathbf{W}_{\mathrm{cls}}^{\mathrm{shadow}} \in \mathbb{R}^{d \times C}$,
initialized as a copy of the base model's classifier head.
The joint classification loss is:
\begin{equation}
  \mathcal{L}_{\mathrm{CLS}} =
    \mathcal{L}_{\mathrm{CE}}\!\left(
      \mathbf{W}_{\mathrm{cls}}\,\mathbf{h}_{\mathrm{base}},\; y
    \right)
    + \lambda\,
    \mathcal{L}_{\mathrm{CE}}\!\left(
      \mathbf{W}_{\mathrm{cls}}^{\mathrm{shadow}}\,\bar{\mathbf{s}}^{(L)},\; y
    \right),
  \label{eq:loss_cls}
\end{equation}
where $y \in \{1,\ldots,C\}$ is the ground-truth class label.

The shadow loss can be interpreted as a regularizer, detailed in the \textcolor{blue}{Appendix \ref{apd:auxiliary_shadow_loss}}.
At inference time, ShadowPEFT supports two deployment configurations: 1) \textbf{\textit{Shadow-attached inference}} with full ShadowPEFT is executed; 
2) \textbf{\textit{Detached shadow-only inference}}, with only the shadow backbone and its prediction head are used. 

These two inference modes highlight a distinctive property of ShadowPEFT: the adaptation module is not merely a set of attached perturbation weights, but a standalone functional component that can be trained jointly with the base model and deployed either together with or separately from it.

\section{Experiments}
\label{sec:experiment}
\subsection{Experimental Setup}
\label{sec:exp_setup}

\begin{table}[htbp!]
\centering
\caption{Generation and understanding results. \textit{ShadowPEFT} is attached inference, and \textit{Detached Shadow Only} evaluates the shadow without the base model. \textbf{Bold} indicates the best result within each block. Reported scores are averages of 5 runs.  ShadowPEFT outperforms baselines significantly on average, with p-value $<5\%$ in paired t-test.}
\label{tab:main_exp}
\setlength{\tabcolsep}{5pt}
\renewcommand{\arraystretch}{1.15}
\resizebox{\textwidth}{!}{%
\begin{tabular}{l r | c c c | c c | c}
\toprule
\multirow{2}{*}{\textbf{Method}} & \multirow{2}{*}{\textbf{Trainable Params}} &
\multicolumn{3}{c|}{\textbf{Generation}} &
\multicolumn{2}{c|}{\textbf{Understanding}} &
\multirow{2}{*}{\textbf{Avg.}} \\
\cmidrule(lr){3-5}\cmidrule(lr){6-7}
& & \textbf{MMLU} & \textbf{GSM8K} & \textbf{SQuAD V2} & \textbf{Amazon} & \textbf{20News} & \\
\midrule

\multicolumn{8}{c}{\textit{Primary compact matched-budget comparisons}} \\
\midrule
\addlinespace[2pt]
\textit{Qwen3 0.6B} & 0 & 37.34 &	47.69	& 49.47 & - & - & - \\
LoRA          & 9.2M  & 49.91          & 48.22          & 80.75          & 60.40          & 69.76          & 61.81 \\
DoRA          & 9.3M  & 50.34          & 48.37          & \textbf{80.91} & 60.92          & 69.88 & 62.08 \\
\rowcolor{gray!20}
ShadowPEFT & 9.1M & \textbf{50.63} & \textbf{48.90} & 80.54 & \textbf{61.18} & \textbf{70.10} & \textbf{62.27} \\
\rowcolor{gray!12}
$\hookrightarrow$ \textrm{Detached Shadow Only}   & 9.1M  & 24.62          & 1.90           & 42.10          & 50.90          & 64.42          & 36.79 \\
\midrule

\textit{Qwen3 4B} & 0 & 67.54 & 	77.18 &	68.48 &	- & 	- &	- \\
LoRA          & 23M & 72.37          & 76.80          & 86.55          & 61.78          & 75.23 & 74.55 \\

DoRA          & 24M & 72.56          & 77.86          & 86.48          & 62.02          & 75.31          & 74.85 \\
\rowcolor{gray!20}
ShadowPEFT & 23M & \textbf{72.91} & \textbf{79.00} & \textbf{86.84} & \textbf{62.66} & \textbf{75.73}          & \textbf{75.43} \\
\rowcolor{gray!12}
$\hookrightarrow$ \textrm{Detached Shadow Only}   & 23M & 27.01          & 1.21           & 41.99          & 51.88          & 69.93          & 38.40 \\
\midrule

\textit{Qwen3 8B} & 0 &	71.92 &	68.84 &	65.98 &	- &	- &	- \\
LoRA          & 31M & 76.46          & 79.76          & 86.90          & 62.42          & \textbf{77.03}          & 76.51 \\

DoRA          & 31M & 75.79          & 78.39          & 86.79          & 62.22          & 76.78          & 75.99 \\
\rowcolor{gray!20}
ShadowPEFT & 29M & \textbf{76.51} & \textbf{80.74} & \textbf{87.51} & \textbf{62.84} & 76.99          & \textbf{76.92} \\
\rowcolor{gray!12}
$\hookrightarrow$ \textrm{Detached Shadow Only}   & 29M & 24.03          & 2.20           & 44.35          & 47.24          & 62.63          & 36.09 \\
\midrule

\multicolumn{8}{c}{\textit{Secondary high-capacity study: Qwen3 8B + roughly 0.5B shadow}} \\
\midrule
\rowcolor{gray!20}
ShadowPEFT w/ random shadow  & 455M & \textbf{76.82} & 80.21 & 87.39 & 62.68 & 75.88 & 76.60 \\
\rowcolor{gray!12}
$\hookrightarrow$ \textrm{Detached Shadow Only}   & 455M & 25.35          & 1.06           & 46.82          & 46.84          & 55.47          & 35.11 \\
\rowcolor{gray!20}
ShadowPEFT w/ pretrained shadow & 455M & 76.54 & \textbf{82.18} & \textbf{87.78} & \textbf{62.72} & \textbf{76.31} & \textbf{77.11} \\
\rowcolor{gray!12}
$\hookrightarrow$ \textrm{Detached Shadow Only}   & 455M & 50.03          & 48.45          & 78.93          & 60.52          & 72.63         & 62.11 \\
\bottomrule
\end{tabular}%
}
\end{table}

\textbf{Models and baselines.}
We conduct experiments primarily on three backbone scales from the Qwen3 family \citep{yang2025qwen3}: 0.6B, 4B, and 8B parameters, comparing against two widely used low-rank PEFT baselines: LoRA \citep{hu2022lora} and DoRA \citep{liu2024dora}. We further extend the backbone to LLaMA~3.2 \citep{grattafiori2024llama} and include DeLoRA \citep{bini2025decoupling}, LoHA \citep{hyeon2021loha}, and LoKr \citep{yeh2024lokr}. However, due to computational constraints, most experiments focus on the two most widely adopted baselines, LoRA and DoRA.

\textbf{Benchmarks.}
We evaluate on five benchmarks including two categories: \textbf{generation and QA-style tasks} (MMLU \citep{hendryckstest2021}, GSM8K \citep{cobbe2021gsm8k}, and SQuAD~v2 \citep{rajpurkar2018squadv2}) and \textbf{understanding tasks} (Amazon review sentiment \citep{keung2020amazonreview} and 20 Newsgroup \citep{twenty_newsgroups} classification). For a fair comparison, we use the same prompt template for each dataset across all methods (details in \textcolor{blue}{Appendix~\ref{apd:prompt_template}}). Beyond these text-based benchmarks, we also test on \textbf{subject-driven image generation} with DreamBooth \citep{ruiz2023dreambooth}.

\textbf{Implementation details.}
We run all experiments on two NVIDIA A800 GPUs.
For LoRA and DoRA, we follow common practice and set the rank $r=32$, scaling factor $\alpha=32$, and dropout rate $0.05$. The learning rate is selected via grid search over the range of $1e-5$ to $1e-3$. 
To maintain similar numbers of trainable parameters across different methods, we modify the number of attention heads and attention layers in the shadow model according to each backbone size. In addition, we adjust the bottleneck dimensions of the shadow injection and update modules.
Full configuration details are provided in the released code repository.

\subsection{Main Results}
\label{sec:main_results}

\textcolor{blue}{Table~\ref{tab:main_exp}} compares performance across three backbone scales. With a similar trainable-parameter budget, ShadowPEFT matches LoRA and DoRA and achieves higher average scores, confirming its effectiveness and robustness. 
We next investigate shadow initialization in a separate high-capacity regime (Qwen3~8B backbone, high capacity 455M \textit{vs.} compact 29M shadow model). 
With random initialization, the attached model obtains an average score of $76.60$, while the detached shadow alone obtains $35.11$. This poor detached result shows that joint task fine-tuning does not by itself produce a capable standalone model in this setting. We then initialize the same high-capacity shadow using a smaller pretrained model, Moore--Penrose pseudo-inverse initialization \citep{barata2012moore}, and additional causal-language-model training (\textcolor{blue}{Appendix~\ref{apd:shadow_pretraining}}). 
The attached average rises to $77.11$; GSM8K rises from $80.21$ to $82.18$, and SQuAD~v2 from $87.39$ to $87.78$. The detached pretrained shadow obtains an average of $62.11$. 
This highlights that detached performance relies heavily on pretrained capability, without isolating pretraining from higher  capacity or proving compact deployment benefits.
Furthermore, we compare ShadowPEFT with more baselines: DeLoRA \citep{bini2025decoupling}, LoHA \citep{hyeon2021loha}, and LoKr \citep{yeh2024lokr}), and more backbones in \textcolor{blue}{Table~\ref{tab:main_more_baselines}}. It shows ShadowPEFT consistently outperforms baselines across datasets and backbones, except for SQuADv2 on Qwen3~0.6B, further validating its effectiveness. In addition, we provide a supplementary system-level edge–cloud evaluation in \textcolor{blue}{Appendix \ref{apd:edge_cloud_study}}.

\begin{table}[t]
\centering
\small
\caption{Comparison of PEFT methods across various backbone models over five runs. For a fair comparison, all models were implemented using the PEFT library under the same settings. }
\label{tab:main_more_baselines}
\setlength{\tabcolsep}{3.5pt}
\resizebox{.95\linewidth}{!}{%
\begin{tabular}{lllccccc}
\toprule
\textbf{Backbone} & \textbf{Method} & \textbf{\#Param} & \textbf{MMLU} & \textbf{GSM8K} & \textbf{SQuAD v2} & \textbf{Amazon} & \textbf{20News} \\
\midrule
\multirow{7}{*}{\shortstack{Qwen3 0.6B \\ \citep{yang2025qwen3}}}
& LoRA \citep{hu2022lora}    & 9.2M & $49.91$ & $48.22$ & $80.75$ & $60.40$ & $69.76$ \\
& DoRA \citep{liu2024dora}   & 9.3M & $50.34$ & $48.37$ & $80.91$ & $60.92$ & $69.88$ \\
& DeLoRA \citep{bini2025decoupling} & 9.2M  & $47.43$ & $48.51$ &  $\textbf{80.92}$ & $60.56$ & $69.34$ \\
& LoHA \citep{hyeon2021loha}  & 9.2M & $47.63$ &$37.91$ &  $78.55$ & $59.22$ & $34.26$ \\
& LoKr \citep{yeh2024lokr} & 9.2M  & $48.20$ & $42.23$ & $79.85$ & $60.00$ & $48.70$ \\
& ShadowPEFT & 9.1M & $\textbf{50.63}$ & $\textbf{48.90}$ & $80.54$ & $\textbf{61.18}$ & $\textbf{70.10}$ \\
\midrule
\multirow{7}{*}{\shortstack{Llama 3.2 1B \\ \citep{grattafiori2024llama}}}
& LoRA \citep{hu2022lora}  &  6.8M & $50.94$ & $40.10$ & $ 81.17$ & $61.48$ & $72.25$ \\
& DoRA \citep{liu2024dora}  & 6.9M  & $51.10$ & $40.11$ & $80.98$ & $61.32$ & $72.29$ \\
& DeLoRA \citep{bini2025decoupling}  & 6.8M & $50.56$ & $41.24$ & $80.77$ & $61.00$ & $73.89$ \\
& LoHA \citep{hyeon2021loha}  & 6.8M & $49.16$ & $37.53$ & $77.42$ & $60.54$ & $49.15$ \\
& LoKr \citep{yeh2024lokr}  & 6.8M & $49.64$ & $39.12$ & $79.33$ & $60.78$ & $61.05$ \\
& ShadowPEFT  & 5.9M & $\textbf{51.23}$ & $\textbf{41.93}$ & $\textbf{81.63}$ & $\textbf{61.90}$ & $\textbf{74.20}$ \\
\bottomrule
\end{tabular}
}
\end{table}

\subsection{Ablation Study}
\label{sec:ablation}
\begin{table}[htbp]
  \centering
  \begin{minipage}[c]{0.52\textwidth}
    The matched-budget PEFT comparison above is separate from this component ablation. Here, we ablate the two crucial modules of ShadowPEFT, Shadow Injection (shadow$\rightarrow$base) and Shadow Update (base$\rightarrow$shadow), on Qwen3~0.6B. Specifically, we evaluate all four on/off combinations on GSM8K (generation) and 20Newsgroups (classification), keeping the shadow model and training recipe fixed. The injection=on and update=on cell is the full ShadowPEFT. \textcolor{blue}{Table \ref{tab:ablation}} presents the results.
  \end{minipage}\hfill
  \begin{minipage}[c]{0.44\textwidth}
    \centering
    \caption{Ablation results of ShadowPEFT. The base model is Qwen3~0.6B.}
    \label{tab:ablation}
    \begin{tabular}{cccc}
    \toprule
    Inject & Update & GSM8K & 20News \\
    \midrule
    \textcolor{red}{off} & \textcolor{red}{off} & $1.74$ & $34.94$ \\
    \textcolor{red}{off}  & \textcolor{ForestGreen}{on} & $1.74$ & $34.94$ \\
    \textcolor{ForestGreen}{on} & \textcolor{red}{off}  & $46.93$ & $69.17$ \\
    \textcolor{ForestGreen}{on}  & \textcolor{ForestGreen}{on}  & $\mathbf{48.90}$ & $\mathbf{70.10}$ \\
    \bottomrule
    \end{tabular}
  \end{minipage}
\vspace{-18px}
\end{table}
\medskip
\noindent
We observe that injection is necessary for adapting the base model: disabling it collapses GSM8K to $1.74$ and 20News to $34.94$, whether or not update is enabled. Holding injection on and removing update causes a drop of $1.97$ points on GSM8K ($48.90\rightarrow46.93$) and $0.93$ on 20News ($70.10\rightarrow69.17$). These results highlight the importance of both injection and update modules.

\subsection{Discussion on Trainable Parameter Scaling}
\label{sec:scaling}
We further study how trainable parameter scale affects ShadowPEFT on Qwen3~8B, comparing it with LoRA and DoRA under comparable parameter budgets in \textcolor{blue}{Figure~\ref{fig:scaling_latency}(a)}. Overall, ShadowPEFT benefits from increasing shadow size and reaches its best performance around the mid-to-large scale, after which the gains begin to saturate. In contrast, LoRA remains relatively flat as the parameter budget increases, while DoRA shows a clear degradation trend at larger scales.
In this sweep, simply increasing low-rank adaptation capacity does not improve performance and can coincide with lower scores, consistent with prior observations on rank-based PEFT~\citep{rathore2025much}. 
ShadowPEFT shows a different scaling curve over the measured parameter range. Performance improves steadily as the number of trainable parameters increases up to 0.4~B, demonstrating the method's robustness to larger capacities. However, beyond this point, from 0.4~B to 0.5~B, performance begins to decline. 
This saturation effect aligns with the observation that a randomly initialized large shadow (0.5~B) does not substantially outperform the compact configuration in \textcolor{blue}{Table \ref{tab:main_exp}}, suggesting diminishing returns and potential overcapacity challenges at very high parameter budgets. We also find that pretraining shadow can alleviate this issue according to \textcolor{blue}{Table \ref{tab:main_exp}}.

\begin{figure}[htbp]
    \centering
    \begin{tikzpicture}[
    font=\fontfamily{ptm}\selectfont,
    ]
        \begin{axis}[
            name=scaling,
            width=0.52\textwidth,
            height=0.35\textwidth,
            xlabel={(a) Parameter Scaling},
            ylabel={GSM8K Accuracy (\%)},
            ylabel style={font=\tiny, yshift=2pt},
            xmin=0.05, xmax=0.55,
            ymin=77.0, ymax=83.2,
            xtick={0.1, 0.2, 0.3, 0.4, 0.5},
            xticklabels={0.1B, 0.2B, 0.3B, 0.4B, 0.5B},
            ytick={78, 79, 80, 81, 82, 83},
            xticklabel style={font=\tiny},
            yticklabel style={font=\tiny},
            grid=both,
            grid style={line width=0.2pt, draw=gray!25},
            major grid style={line width=0.2pt, draw=gray!40},
            minor tick num=1,
            axis line style={gray!60},
            tick style={gray!60},
            legend style={
                font=\tiny,
                at={(0.03, 0.06)},
                anchor=south west,
                draw=gray!50,
                fill=white,
                fill opacity=0.9,
                text opacity=1,
                inner sep=4pt,
                row sep=1pt,
                legend cell align=left,
            },
            tick align=outside,
            clip=false,
        ]

        \addplot[
            color=RoyalBlue,
            dashed,
            line width=1.2pt,
            mark=o,
            mark size=2.8pt,
            mark options={solid, fill=RoyalBlue!30, draw=RoyalBlue, line width=1pt},
        ] coordinates {
            (0.1, 80.67) (0.2, 81.28) (0.3, 80.82) (0.4, 81.05) (0.5, 80.52)
        };
        \addlegendentry{LoRA}

        \addplot[
            color=BrickRed,
            dashed,
            line width=1.2pt,
            mark=square,
            mark size=2.6pt,
            mark options={solid, fill=BrickRed!25, draw=BrickRed, line width=1pt},
        ] coordinates {
            (0.1, 81.12) (0.2, 80.36) (0.3, 79.76) (0.4, 79.08) (0.5, 77.79)
        };
        \addlegendentry{DoRA}

        \addplot[
            color=ForestGreen,
            solid,
            line width=1.8pt,
            mark=diamond*,
            mark size=3.2pt,
            mark options={solid, fill=ForestGreen!50, draw=ForestGreen, line width=1pt},
        ] coordinates {
            (0.1, 81.35) (0.2, 81.50) (0.3, 81.96) (0.4, 82.12) (0.5, 81.80)
        };
        \addlegendentry{ShadowPEFT}

        \end{axis}

        \begin{axis}[
            name=latency,
            at={(scaling.south east)},
            anchor=south west,
            xshift=16pt,
            width=0.47\textwidth,
            height=0.31\textwidth,
            ybar,
            bar width=8pt,
            xlabel={(b) Inference Latency},
            ylabel={Mean Latency (ms)},
            ylabel style={font=\tiny, rotate=-180, anchor=south},
            ymin=0, ymax=185,
            ytick={0, 40, 80, 120, 160},
            yticklabel pos=right,
            yticklabel style={font=\tiny},
            ytick pos=right,
            xtick=data,
            xticklabels={Qwen3 0.6B, Qwen3 4B, Qwen3 8B},
            xticklabel style={font=\tiny},
            enlarge x limits=0.28,
            grid= both,
            grid style={line width=0.3pt, draw=gray!30},
            axis line style={gray!60},
            tick style={gray!60},
            tick align=outside,
            legend style={
                font=\tiny,
                at={(0.5, 1.06)},
                anchor=south,
                draw=gray!50,
                fill=white,
                inner sep=3pt,
                column sep=4.5pt,
                legend columns=3,
                legend cell align=left,
            },
            clip=false,
            title={\tiny (b) Inference Latency},
            title style={yshift=-2pt},
        ]

        \addplot[
            fill=RoyalBlue!55,
            draw=RoyalBlue!80,
            line width=0.6pt,
            nodes near coords,
            nodes near coords style={
                font=\tiny,
                /pgf/number format/fixed,
                /pgf/number format/precision=1,
                anchor=south,
                yshift=1pt,
                text=black!70,
            },
            error bars/.cd,
                y dir=both, y explicit,
                error bar style={line width=0.8pt, draw=RoyalBlue!80},
                error mark options={rotate=90, mark size=2pt, line width=0.8pt},
        ] coordinates {
            (1, 80.996)  +- (0, 1.063)
            (2, 101.157) +- (0, 0.221)
            (3, 103.341) +- (0, 1.018)
        };
        \addlegendentry{LoRA}

        \addplot[
            fill=BrickRed!50,
            draw=BrickRed!75,
            line width=0.6pt,
            nodes near coords,
            nodes near coords style={
                font=\tiny,
                /pgf/number format/fixed,
                /pgf/number format/precision=1,
                anchor=south,
                yshift=1pt,
                text=black!70,
            },
            error bars/.cd,
                y dir=both, y explicit,
                error bar style={line width=0.8pt, draw=BrickRed!80},
                error mark options={rotate=90, mark size=2pt, line width=0.8pt},
        ] coordinates {
            (1, 121.544) +- (0, 1.396)
            (2, 155.982) +- (0, 0.947)
            (3, 152.664) +- (0, 1.110)
        };
        \addlegendentry{DoRA}

        \addplot[
            fill=ForestGreen!50,
            draw=ForestGreen!75,
            line width=0.6pt,
            nodes near coords,
            nodes near coords style={
                font=\tiny,
                /pgf/number format/fixed,
                /pgf/number format/precision=1,
                anchor=south,
                yshift=1pt,
                text=black!70,
            },
            error bars/.cd,
                y dir=both, y explicit,
                error bar style={line width=0.8pt, draw=ForestGreen!80},
                error mark options={rotate=90, mark size=2pt, line width=0.8pt},
        ] coordinates {
            (1, 84.015)  +- (0, 0.729)
            (2, 107.172) +- (0, 0.347)
            (3, 109.194) +- (0, 0.995)
        };
        \addlegendentry{ShadowPEFT}

        \end{axis}
    \end{tikzpicture}
    \caption{%
        \textbf{(a) Parameter scaling:} GSM8K accuracy of LoRA, DoRA, and ShadowPEFT with the trainable parameter scales from 0.1B to 0.5B (base model: Qwen3 8B).
        \textbf{(b) Inference latency} (mean $\pm$ std over 10 attempts) across three base model sizes.
    }
    \label{fig:scaling_latency}
\end{figure}

\subsection{Discussion on Generalization Performance}
\label{sec:generalization}
\begin{table*}[!htbp]
    \vspace{-4pt}
    \centering
    \begin{minipage}[t]{0.45\textwidth}
        \vspace{0pt}
        We evaluate out-of-distribution (OOD) generalization by fine-tuning Qwen3~4B on one dataset and testing on the remaining generation benchmarks with 2-shot demonstrations, as shown in \textcolor{blue}{Table~\ref{tab:generalization}}. ShadowPEFT achieves the best OOD average across all training settings, consistently outperforming LoRA and DoRA.
        These results indicate that ShadowPEFT improves downstream adaptation without sacrificing transferability. Its coordinated refinement appears to preserve generalization better than distributed weight-level updates, making it suitable for practical scenarios where the deployment distribution may differ from the fine-tuning data.
    \end{minipage}
    \hfill
    \begin{minipage}[t]{0.53\textwidth}
        \vspace{0pt}
        \centering
        \caption{Out-of-distribution generalization results. \colorbox{cyan!20}{Blue} cells indicate in-distribution evaluation. \textbf{Bold} indicates the best result per training group. 
        }
        \label{tab:generalization}
        \scriptsize
        \setlength{\tabcolsep}{3.2pt}
        \renewcommand{\arraystretch}{1.08}
        \begin{tabular}{@{}llccccc@{}}
        \toprule
        \textbf{Method} & \textbf{Train} & \textbf{GSM8K} & \textbf{SQuAD} & \textbf{MMLU} & \textbf{Avg.} & \textbf{OOD Avg.} \\
        \midrule
        LoRA & \multirow{3}{*}{GSM8K}
            & \cellcolor{cyan!20}72.37 & 35.96 & 64.83 & 57.72 & 50.40 \\
        DoRA &
            & \cellcolor{cyan!20}77.86 & 31.70 & \textbf{65.43} & 58.33 & 48.57 \\
        \rowcolor{gray!15}
        Ours &
            & \cellcolor{cyan!20}\textbf{79.00} & \textbf{38.84} & 62.37 & \textbf{60.07} & \textbf{50.61} \\
        \midrule
        LoRA & \multirow{3}{*}{SQuAD}
            & 41.12 & \cellcolor{cyan!20}86.55 & \textbf{63.70} & 63.79 & 52.41 \\
        DoRA &
            & 42.64 & \cellcolor{cyan!20}86.48 & 63.20 & 64.11 & 52.92 \\
        \rowcolor{gray!15}
        Ours &
            & \textbf{42.76} & \cellcolor{cyan!20}\textbf{86.84} & 63.69 & \textbf{64.43} & \textbf{53.23} \\
        \midrule
        LoRA & \multirow{3}{*}{MMLU}
            & \textbf{87.49} & 65.65 & \cellcolor{cyan!20}72.37 & 75.17 & 76.57 \\
        DoRA &
            & 86.58 & 63.77 & \cellcolor{cyan!20}72.56 & 74.30 & 75.18 \\
        \rowcolor{gray!15}
        Ours &
            & 87.41 & \textbf{65.87} & \cellcolor{cyan!20}\textbf{72.91} & \textbf{75.40} & \textbf{76.64} \\
        \bottomrule
        \end{tabular}
    \end{minipage}
    \vspace{-4pt}
\end{table*}

\subsection{Discussion on Efficiency}
\label{sec:latency}

\textcolor{blue}{Figure~\ref{fig:scaling_latency}(b)} shows inference latency (mean ± std, 10 runs) for LoRA, DoRA, and ShadowPEFT across three model sizes. Compared to LoRA, ShadowPEFT adds ~4–6\% latency (e.g., +3.0 ms on Qwen3~0.6B, +6.0 ms on Qwen3~4B, +5.9 ms on Qwen3~8B). DoRA incurs even higher latency. 
\textcolor{blue}{Table~\ref{tab:efficiency}} extends the comparison to additional PEFT methods. ShadowPEFT maintains near-baseline LoRA throughput and significantly outperforms methods like DoRA and LoHA in both prefill and decoding speed. While the dual-path design introduces a modest increase in peak GPU memory, this trade-off effectively preserves high-speed token generation while remaining fully compatible with standard KV-caching optimizations.

\vspace{-5px}
\begin{table}[t]
\centering
\caption{Inference throughput and peak memory on Qwen3~0.6B (GSM8K). Measured with a 128-token prompt and 128 generated tokens. 
Peak memory is max GPU allocation.}
\label{tab:efficiency}
\resizebox{.9\textwidth}{!}{
\begin{tabular}{lccccc}
\toprule
\textbf{Method} & \textbf{KV Cache} & \textbf{Prefill tokens / s} & \textbf{Decode tokens / s} & \textbf{Peak memory (MiB)} \\
\midrule
LoRA \citep{hu2022lora}       & \textcolor{ForestGreen}{on} / \textcolor{red}{off}  & \textcolor{ForestGreen}{$5991$} / \textcolor{red}{$6219$} & \textcolor{ForestGreen}{$50.8$} / \textcolor{red}{$48.2$} & \textcolor{ForestGreen}{$1215$} / \textcolor{red}{$1311$} \\
DoRA \citep{liu2024dora}      & \textcolor{ForestGreen}{on} / \textcolor{red}{off}  & \textcolor{ForestGreen}{$3609$} / \textcolor{red}{$3663$} & \textcolor{ForestGreen}{$29.3$} / \textcolor{red}{$28.5$} & \textcolor{ForestGreen}{$1227$} / \textcolor{red}{$1312$} \\
DeLoRA \citep{bini2025decoupling}   & \textcolor{ForestGreen}{on} / \textcolor{red}{off}  & \textcolor{ForestGreen}{$4171$} / \textcolor{red}{$4251$} & \textcolor{ForestGreen}{$34.6$} / \textcolor{red}{$33.0$} & \textcolor{ForestGreen}{$1215$} / \textcolor{red}{$1312$} \\
LoHA \citep{hyeon2021loha}   & \textcolor{ForestGreen}{on} / \textcolor{red}{off}  & \textcolor{ForestGreen}{$4954$} / \textcolor{red}{$5115$} & \textcolor{ForestGreen}{$39.9$} / \textcolor{red}{$39.7$} & \textcolor{ForestGreen}{$1227$} / \textcolor{red}{$1311$} \\
LoKr \citep{yeh2024lokr}   & \textcolor{ForestGreen}{on} / \textcolor{red}{off}  & \textcolor{ForestGreen}{$5665$} / \textcolor{red}{$5855$} & \textcolor{ForestGreen}{$46.2$} / \textcolor{red}{$45.5$}  & \textcolor{ForestGreen}{$1224$} / \textcolor{red}{$1311$} \\
ShadowPEFT & \textcolor{ForestGreen}{on} / \textcolor{red}{off} & \textcolor{ForestGreen}{$5637$} / \textcolor{red}{$5819$} & \textcolor{ForestGreen}{$47.5$} / \textcolor{red}{$44.9$} & \textcolor{ForestGreen}{$1550$} / \textcolor{red}{$1656$} \\
\bottomrule
\vspace{-30px}
\end{tabular}}
\end{table}

\subsection{Discussion on Subject-driven Image Generation}
\label{sec:dino}

Although the preceding experiments focus primarily on text-based tasks, we further examine ShadowPEFT on image generation. Specifically, we consider subject-driven generation under the DreamBooth setting \citep{ruiz2023dreambooth}, in which Stable Diffusion \citep{rombach2022high} is adapted to generate a reference subject in contexts specified by text prompts. Full experimental details are provided in \textcolor{blue}{Appendix~\ref{apd:dino}}. As shown in \textcolor{blue}{Table~\ref{tab:dino}}, ShadowPEFT achieves a DINO score of $0.693$, matching the strongest hyperparameter-tuned baseline, DeLoRA, while obtaining the highest CLIP-I score of $0.823$. It also requires slightly fewer trainable parameters than the rank-16 low-rank baselines. These results demonstrate that ShadowPEFT effectively preserves subject identity and maintains strong visual similarity to the reference images.

\begin{figure}[htbp]
\centering
\includegraphics[width=\textwidth]
{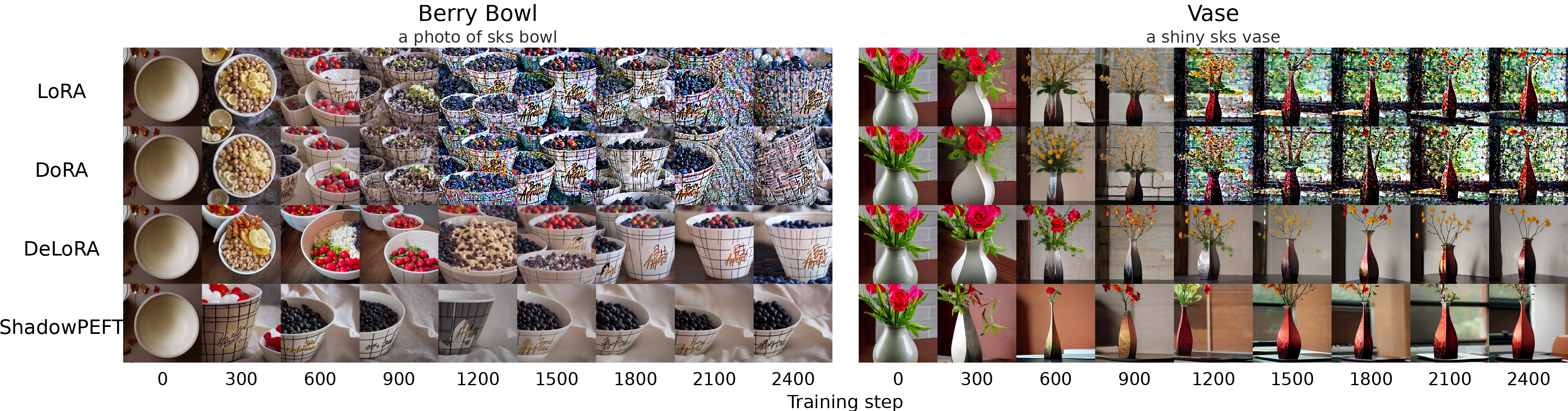}
\caption{Training progression on the \texttt{berry\_bowl} (left) and \texttt{vase} (right) subjects. The special identifier \texttt{sks} denotes the specific subject learned from the reference images. Columns show training steps; rows compare PEFT methods using the displayed prompt and the same random seed.}
\label{fig:dino-training-progression}
\end{figure}

\vspace{-5px}
\begin{minipage}[t]{0.42\textwidth}
    \vspace{0pt}
    \centering
    \captionof{table}{Results of subject-driven image generation.
    $\dagger$ and $\ddagger$ denote results from \citep{ruiz2023dreambooth} and 
    \citep{bini2025decoupling}, respectively. 
    }
    \label{tab:dino}
    \resizebox{\linewidth}{!}{%
    \begin{tabular}{lccc}
        \toprule
        Method & \#Param & DINO & CLIP-I \\
        \midrule
        Real Images & N/A & $0.703$ & $0.864$ \\
        \midrule
        DreamBooth$^\dagger$ & 859.5M & $0.644$ & $0.793$ \\
        LoRA$_{r=16}^\ddagger$ & 3.2M & $0.688$ & $0.818$ \\
        DoRA$_{r=16}^\ddagger$ & 3.2M & $0.689$ & $0.819$ \\
        DeLoRA$_{r=16}^\ddagger$ & 3.2M & $\mathbf{0.693}$ & $0.820$ \\
        \midrule
        ShadowPEFT & 3.1M & $\mathbf{0.693}$ & $\mathbf{0.823}$ \\
        \bottomrule
    \end{tabular}%
    }
\end{minipage}
\hfill
\begin{minipage}[t]{0.54\textwidth}
    \vspace{0pt}
    \small
    \textcolor{blue}{Figure~\ref{fig:dino-training-progression}} provides a qualitative view of subject fidelity and training stability. For the berry bowl, LoRA and DoRA learn the subject at early checkpoints but eventually develop severe high-frequency artifacts. DeLoRA remains stable, but its generations show less consistent contents and weaker subject-specific details. In contrast, ShadowPEFT preserves the characteristic blueberry-filled bowl while maintaining realistic image quality. The vase results exhibit a similar pattern. LoRA and DoRA become corrupted at later checkpoints, whereas DeLoRA remains coherent but gradually deviates from the specific requirement. ShadowPEFT better preserves the tapered silhouette, red decoration, and characteristic flower arrangement while also satisfying the ``shiny'' prompt. Overall, ShadowPEFT yields training stability with stronger preservation of subject-specific features.
\end{minipage}


\section{Conclusion}


We have presented ShadowPEFT, a modular PEFT framework that introduces an explicit depth-wise state for coordinating task adaptation across layers. 
Extensive experiments demonstrate that ShadowPEFT achieves competitive performance against LoRA and DoRA under comparable trainable-parameter budgets. 
Beyond raw accuracy, the framework offers unique deployment advantages: the shadow model can be detached and used independently for inference, supporting edge–cloud collaboration and other resource-constrained scenarios. 
Despite these benefits, ShadowPEFT introduces additional computational overhead during training due to the forward pass through both the backbone and the shadow model. 
Future work may explore more efficient shadow architectures. We believe ShadowPEFT opens a promising direction for rethinking how task-specific knowledge is structured and propagated in large-scale model adaptation.

\section*{Limitations}
\label{sec:limitation}
We have identified several limitations from current model design. First, unlike LoRA and its variants, whose trained weights can be easily merged with those of the base model, ShadowPEFT's shadow model cannot be merged into the base model directly, which may complicate deployment in settings that require a single unified checkpoint. 
Second, the dual-path design of ShadowPEFT incurs a modest increase in peak GPU memory usage due to the dual KV-cache. While this trade-off preserves high-speed token generation and remains fully compatible with standard KV-caching optimizations, it could be a concern for extremely resource-constrained environments. 
Future work will explore more efficient architectural designs to reduce the memory overhead and investigate alternative strategies for merging or distilling the shadow model into the base model.

\subsection*{AI use statement}
In this work, we used generative AI tools for enhancing the clarity, readability, grammar, and overall presentation of the writing. We have not used generative AI tools for generating, manipulating, or fabricating data, results, experiments, or research findings, and any other disclosure categories not relevant to writing assistance are not applicable to this work. Additionally, we used generative AI tools for limited editorial support, such as improving phrasing, sentence structure, and style.

We have reviewed all AI-assisted content to ensure accuracy, originality, and consistency with the authors’ intended meaning. All substantive scientific contributions, including the study design, methodology, analysis, interpretation, and conclusions, were developed and verified by the authors. We take full responsibility for the final content of this work, including any text produced with the aid of generative AI.

\bibliography{iclr2027_conference}

@inproceedings{houlsby2019parameter,
  title={Parameter-efficient transfer learning for NLP},
  author={Houlsby, Neil and Giurgiu, Andrei and Jastrzebski, Stanislaw and Morrone, Bruna and De Laroussilhe, Quentin and Gesmundo, Andrea and Attariyan, Mona and Gelly, Sylvain},
  booktitle={International conference on machine learning},
  pages={2790--2799},
  year={2019},
  organization={PMLR}
}

@inproceedings{li2021prefix,
  title={Prefix-tuning: Optimizing continuous prompts for generation},
  author={Li, Xiang Lisa and Liang, Percy},
  booktitle={Proceedings of the 59th Annual Meeting of the Association for Computational Linguistics and the 11th International Joint Conference on Natural Language Processing (Volume 1: Long Papers)},
  pages={4582--4597},
  year={2021}
}

@article{hu2022lora,
  title={Lora: Low-rank adaptation of large language models.},
  author={Hu, Edward J and Shen, Yelong and Wallis, Phillip and Allen-Zhu, Zeyuan and Li, Yuanzhi and Wang, Shean and Wang, Liang and Chen, Weizhu and others},
  journal={Iclr},
  volume={1},
  number={2},
  pages={3},
  year={2022}
}

@inproceedings{liu2024dora,
  title={Dora: Weight-decomposed low-rank adaptation},
  author={Liu, Shih-Yang and Wang, Chien-Yi and Yin, Hongxu and Molchanov, Pavlo and Wang, Yu-Chiang Frank and Cheng, Kwang-Ting and Chen, Min-Hung},
  booktitle={Forty-first International Conference on Machine Learning},
  year={2024}
}

@article{dettmers2023qlora,
  title={Qlora: Efficient finetuning of quantized llms},
  author={Dettmers, Tim and Pagnoni, Artidoro and Holtzman, Ari and Zettlemoyer, Luke},
  journal={Advances in neural information processing systems},
  volume={36},
  pages={10088--10115},
  year={2023}
}

@article{cobbe2021gsm8k,
  title={Training Verifiers to Solve Math Word Problems},
  author={Cobbe, Karl and Kosaraju, Vineet and Bavarian, Mohammad and Chen, Mark and Jun, Heewoo and Kaiser, Lukasz and Plappert, Matthias and Tworek, Jerry and Hilton, Jacob and Nakano, Reiichiro and Hesse, Christopher and Schulman, John},
  journal={arXiv preprint arXiv:2110.14168},
  year={2021}
}

@inproceedings{rajpurkar2018squadv2,
  title={Know what you don’t know: Unanswerable questions for SQuAD},
  author={Rajpurkar, Pranav and Jia, Robin and Liang, Percy},
  booktitle={Proceedings of the 56th Annual Meeting of the Association for Computational Linguistics (Volume 2: Short Papers)},
  pages={784--789},
  year={2018}
}

@inproceedings{keung2020amazonreview,
  title={The multilingual amazon reviews corpus},
  author={Keung, Phillip and Lu, Yichao and Szarvas, Gy{\"o}rgy and Smith, Noah A},
  booktitle={Proceedings of the 2020 conference on empirical methods in natural language processing (EMNLP)},
  pages={4563--4568},
  year={2020}
}

@misc{lozhkov2024fineweb-edu,
    author       = { Lozhkov, Anton and Ben Allal, Loubna and von Werra, Leandro and Wolf, Thomas },  
    title        = { FineWeb-Edu: the Finest Collection of Educational Content }, 
    year         = 2024,  
    url          = { https://huggingface.co/datasets/HuggingFaceFW/fineweb-edu },  
    doi          = { 10.57967/hf/2497 },
    publisher    = { Hugging Face }
}

@misc{wudao,
  author       = {Zhao Xue and Hanyu Zhao and Sha Yuan and Yequan Wang},
  title        = {{WuDaoCorpora Text}},
  year         = 2022,
  month        = dec,
  publisher    = {Science Data Bank},
  version      = {V1},
  doi          = {10.57760/sciencedb.o00126.00004}, }

@inproceedings{rathore2025much,
  title={How Much is Too Much? Exploring LoRA Rank Trade-offs for Retaining Knowledge and Domain Robustness},
  author={Rathore, Darshita and Kumar, Vineet and Bansal, Chetna and Moitra, Anindya},
  booktitle={Proceedings of the 14th International Joint Conference on Natural Language Processing and the 4th Conference of the Asia-Pacific Chapter of the Association for Computational Linguistics},
  pages={1003--1013},
  year={2025}
}

@article{barata2012moore,
  title={The Moore--Penrose pseudoinverse: A tutorial review of the theory},
  author={Barata, Jo{\~a}o Carlos Alves and Hussein, Mahir Saleh},
  journal={Brazilian Journal of Physics},
  volume={42},
  number={1},
  pages={146--165},
  year={2012},
  publisher={Springer}
}

@misc{twenty_newsgroups,
  author       = {Mitchell, Tom},
  title        = {{Twenty Newsgroups}},
  year         = {1997},
  howpublished = {UCI Machine Learning Repository},
  note         = {{DOI}: https://doi.org/10.24432/C5C323}
}

@article{hendryckstest2021,
  title={Measuring Massive Multitask Language Understanding},
  author={Dan Hendrycks and Collin Burns and Steven Basart and Andy Zou and Mantas Mazeika and Dawn Song and Jacob Steinhardt},
  journal={Proceedings of the International Conference on Learning Representations (ICLR)},
  year={2021}
}

@inproceedings{lester2021power,
  title={The power of scale for parameter-efficient prompt tuning},
  author={Lester, Brian and Al-Rfou, Rami and Constant, Noah},
  booktitle={Proceedings of the 2021 conference on empirical methods in natural language processing},
  pages={3045--3059},
  year={2021}
}

@inproceedings{liu2022p,
  title={P-tuning: Prompt tuning can be comparable to fine-tuning across scales and tasks},
  author={Liu, Xiao and Ji, Kaixuan and Fu, Yicheng and Tam, Weng and Du, Zhengxiao and Yang, Zhilin and Tang, Jie},
  booktitle={Proceedings of the 60th Annual Meeting of the Association for Computational Linguistics (Volume 2: Short Papers)},
  pages={61--68},
  year={2022}
}

@inproceedings{chronopoulou2023adaptersoup,
  title={Adaptersoup: Weight averaging to improve generalization of pretrained language models},
  author={Chronopoulou, Alexandra and Peters, Matthew E and Fraser, Alexander and Dodge, Jesse},
  booktitle={Findings of the Association for Computational Linguistics: EACL 2023},
  pages={2054--2063},
  year={2023}
}

@inproceedings{zhao2022tiny,
  title={Tiny-attention adapter: Contexts are more important than the number of parameters},
  author={Zhao, Hongyu and Tan, Hao and Mei, Hongyuan},
  booktitle={Proceedings of the 2022 Conference on Empirical Methods in Natural Language Processing},
  pages={6626--6638},
  year={2022}
}

@article{karimi2021compacter,
  title={Compacter: Efficient low-rank hypercomplex adapter layers},
  author={Karimi Mahabadi, Rabeeh and Henderson, James and Ruder, Sebastian},
  journal={Advances in neural information processing systems},
  volume={34},
  pages={1022--1035},
  year={2021}
}

@inproceedings{zhangadaptive,
  title={Adaptive Budget Allocation for Parameter-Efficient Fine-Tuning},
  author={Zhang, Qingru and Chen, Minshuo and Bukharin, Alexander and He, Pengcheng and Cheng, Yu and Chen, Weizhu and Zhao, Tuo},
  booktitle={The Eleventh International Conference on Learning Representations},
  year={2023}
}

@inproceedings{valipour2023dylora,
  title={DyLoRA: Parameter-efficient tuning of pre-trained models using dynamic search-free low-rank adaptation},
  author={Valipour, Mojtaba and Rezagholizadeh, Mehdi and Kobyzev, Ivan and Ghodsi, Ali},
  booktitle={Proceedings of the 17th Conference of the European Chapter of the Association for Computational Linguistics},
  pages={3274--3287},
  year={2023}
}

@inproceedings{xu2023qa,
  title={QA-LoRA: Quantization-Aware Low-Rank Adaptation of Large Language Models},
  author={Xu, Yuhui and Xie, Lingxi and Gu, Xiaotao and Chen, Xin and Chang, Heng and Zhang, Hengheng and Chen, Zhengsu and Zhang, Xiaopeng and Tian, Qi},
  booktitle={The Twelfth International Conference on Learning Representations},
  year={2023}
}

@article{wang2024lora,
  title={Lora-ga: Low-rank adaptation with gradient approximation},
  author={Wang, Shaowen and Yu, Linxi and Li, Jian},
  journal={Advances in Neural Information Processing Systems},
  volume={37},
  pages={54905--54931},
  year={2024}
}

@article{li2024mixlora,
  title={Mixlora: Enhancing large language models fine-tuning with lora-based mixture of experts},
  author={Li, Dengchun and Ma, Yingzi and Wang, Naizheng and Ye, Zhengmao and Cheng, Zhiyuan and Tang, Yinghao and Zhang, Yan and Duan, Lei and Zuo, Jie and Yang, Cal and others},
  journal={arXiv preprint arXiv:2404.15159},
  year={2024}
}

@inproceedings{che2026lora,
  title={LoRA in LoRA: Towards parameter-efficient architecture expansion for continual visual instruction tuning},
  author={Che, Chang and Wang, Ziqi and Yang, Pengwan and Wang, Cheems and Ma, Hui and Shi, Zenglin},
  booktitle={Proceedings of the AAAI Conference on Artificial Intelligence},
  volume={40},
  number={24},
  pages={19978--19986},
  year={2026}
}

@article{hsu2024safe,
  title={Safe lora: The silver lining of reducing safety risks when finetuning large language models},
  author={Hsu, Chia-Yi and Tsai, Yu-Lin and Lin, Chih-Hsun and Chen, Pin-Yu and Yu, Chia-Mu and Huang, Chun-Ying},
  journal={Advances in Neural Information Processing Systems},
  volume={37},
  pages={65072--65094},
  year={2024}
}

@inproceedings{yang2025mtl,
  title={Mtl-lora: Low-rank adaptation for multi-task learning},
  author={Yang, Yaming and Muhtar, Dilxat and Shen, Yelong and Zhan, Yuefeng and Liu, Jianfeng and Wang, Yujing and Sun, Hao and Deng, Weiwei and Sun, Feng and Zhang, Qi and others},
  booktitle={Proceedings of the AAAI Conference on Artificial intelligence},
  volume={39},
  number={20},
  pages={22010--22018},
  year={2025}
}

@article{li2026ls,
  title={LS-BiLLMs: Label supervised bi-directional large language models for token-and sequence-level information extraction},
  author={Li, Zongxi and Li, Xianming and Li, Jing and Xie, Haoran and Wang, Fu Lee and Li, Qing},
  journal={Information Processing \& Management},
  volume={63},
  number={4},
  pages={104568},
  year={2026},
  publisher={Elsevier}
}

@article{Han2024ParameterEfficientFF,
  title={Parameter-Efficient Fine-Tuning for Large Models: A Comprehensive Survey},
  author={Zeyu Han and Chao Gao and Jinyang Liu and Jeff Zhang and Sai Qian Zhang},
  journal={Trans. Mach. Learn. Res.},
  year={2024},
  volume={2024},
}

@inproceedings{Cho2014GRU,
  title={Learning Phrase Representations using RNN Encoder–Decoder for Statistical Machine Translation},
  author={Kyunghyun Cho and Bart van Merrienboer and Çaglar G{\"u}lçehre and Dzmitry Bahdanau and Fethi Bougares and Holger Schwenk and Yoshua Bengio},
  booktitle={Conference on Empirical Methods in Natural Language Processing},
  year={2014},
}

@inproceedings{zhang2020sidetuning,
  title={Side-tuning: a baseline for network adaptation via additive side networks},
  author={Zhang, Jeffrey O and Sax, Alexander and Zamir, Amir and Guibas, Leonidas and Malik, Jitendra},
  booktitle={European conference on computer vision},
  pages={698--714},
  year={2020},
  organization={Springer}
}

@inproceedings{sung2022lst,
  title={Lst: Ladder side-tuning for parameter and memory efficient transfer learning},
  author={Sung, Yi-Lin and Cho, Jaemin and Bansal, Mohit},
  journal={Advances in Neural Information Processing Systems},
  volume={35},
  pages={12991--13005},
  year={2022}
}

@article{buyukakyuz2024olora,
  title={{OLoRA}: Orthonormal Low-Rank Adaptation of Large Language Models},
  author={B{\"u}y{\"u}kaky{\"u}z, Kerim},
  journal={arXiv preprint arXiv:2406.01775},
  year={2024},
  doi={10.48550/arXiv.2406.01775}
}

@inproceedings{meng2024pissa,
  title={{PiSSA}: Principal Singular Values and Singular Vectors Adaptation of Large Language Models},
  author={Meng, Fanxu and Wang, Zhaohui and Zhang, Muhan},
  booktitle={Advances in Neural Information Processing Systems},
  volume={37},
  year={2024},
  doi={10.48550/arXiv.2404.02948}
}

@inproceedings{bini2025decoupling,
  title={Decoupling angles and strength in low-rank adaptation},
  author={Bini, Massimo and Girrbach, Leander and Akata, Zeynep},
  booktitle={International Conference on Learning Representations},
  volume={2025},
  pages={20216--20233},
  year={2025}
}

@inproceedings{ruiz2023dreambooth,
  title={Dreambooth: Fine tuning text-to-image diffusion models for subject-driven generation},
  author={Ruiz, Nataniel and Li, Yuanzhen and Jampani, Varun and Pritch, Yael and Rubinstein, Michael and Aberman, Kfir},
  booktitle={2023 IEEE/CVF Conference on Computer Vision and Pattern Recognition (CVPR)},
  pages={22500--22510},
  year={2023},
  organization={IEEE}
}

@inproceedings{rombach2022high,
  title={High-resolution image synthesis with latent diffusion models},
  author={Rombach, Robin and Blattmann, Andreas and Lorenz, Dominik and Esser, Patrick and Ommer, Bj{\"o}rn},
  booktitle={2022 IEEE/CVF conference on computer vision and pattern recognition (CVPR)},
  pages={10674--10685},
  year={2022},
  organization={ieee}
}

@inproceedings{caron2021emerging,
  title={Emerging properties in self-supervised vision transformers},
  author={Caron, Mathilde and Touvron, Hugo and Misra, Ishan and J{\'e}gou, Herv{\'e} and Mairal, Julien and Bojanowski, Piotr and Joulin, Armand},
  booktitle={2021 IEEE/CVF international conference on computer vision (ICCV)},
  pages={9630--9640},
  year={2021},
  organization={IEEE}
}

@inproceedings{radford2021learning,
  title={Learning transferable visual models from natural language supervision},
  author={Radford, Alec and Kim, Jong Wook and Hallacy, Chris and Ramesh, Aditya and Goh, Gabriel and Agarwal, Sandhini and Sastry, Girish and Askell, Amanda and Mishkin, Pamela and Clark, Jack and others},
  booktitle={International conference on machine learning},
  pages={8748--8763},
  year={2021},
  organization={PmLR}
}

@article{grattafiori2024llama,
  title={The llama 3 herd of models},
  author={Grattafiori, Aaron and Dubey, Abhimanyu and Jauhri, Abhinav and Pandey, Abhinav and Kadian, Abhishek and Al-Dahle, Ahmad and Letman, Aiesha and Mathur, Akhil and Schelten, Alan and Vaughan, Alex and others},
  journal={arXiv preprint arXiv:2407.21783},
  year={2024}
}

@article{yang2025qwen3,
  title={Qwen3 technical report},
  author={Yang, An and Li, Anfeng and Yang, Baosong and Zhang, Beichen and Hui, Binyuan and Zheng, Bo and Yu, Bowen and Gao, Chang and Huang, Chengen and Lv, Chenxu and others},
  journal={arXiv preprint arXiv:2505.09388},
  year={2025}
}

@article{hyeon2021loha,
  title={Fedpara: Low-rank hadamard product for communication-efficient federated learning},
  author={Nam, Hyeon-Woo and Moon, Ye-Bin and Oh, Tae-Hyun},
  journal={arXiv preprint arXiv:2108.06098},
  year={2021}
}

@inproceedings{yeh2024lokr,
  title={Navigating text-to-image customization: From lycoris fine-tuning to model evaluation},
  author={Yeh, Shih-Ying and Hsieh, Yu-Guan and Gao, Zhidong and Yang, Bernard and Oh, Giyeong and Gong, Yanmin},
  booktitle={International Conference on Learning Representations},
  volume={2024},
  pages={27099--27110},
  year={2024}
}
\bibliographystyle{iclr2027_conference}

\appendix
\section{Shadow Backbone Designs}
\label{apd:shadow_backbone_detail}

\paragraph{Implicit vs.\ explicit shadow models.}
ShadowPEFT supports two shadow-backbone initialization strategies.
In the \emph{implicit} setting the shadow model is derived automatically from the base model's configuration by reducing the number of layers to
$L_s \ll L$ and optionally narrowing the intermediate width and attention heads.
In the \emph{explicit} setting the practitioner provides an independent shadow model of any compatible architecture, enabling cross-model knowledge transfer.

\paragraph{Embedding sharing.}
To avoid maintaining a duplicate token embedding lookup table inside the shadow model, ShadowPEFT shares the base model's frozen embedding layer.
Input token embeddings are computed once as $\mathbf{E} = \texttt{Embed}_{\text{base}}(\mathbf{x}) \in \mathbb{R}^{B \times T \times d}$ and fed to the shadow model as \texttt{inputs\_embeds}, so the shadow \texttt{embed\_tokens} matrix is removed from the module entirely.

\paragraph{Hidden-size projection.}
When the shadow model has a hidden dimension $d_s \neq d$, a learned linear projection
$\mathbf{W}_{\text{proj}} \in \mathbb{R}^{d_s \times d}$ (no bias) is applied to align
the shadow output with the representation space of the base model:
\begin{equation}
  \mathbf{s}^{(0)} \leftarrow \mathbf{s}^{(0)}\,\mathbf{W}_{\text{proj}}.
\end{equation}
When $d_s = d$, the projection is an identity and does not introduce additional parameters.

\section{Gated Residual Update}
\label{apd:gated_residual_update}

In practice, both $T^{(\ell)}(\cdot)$ and $G^{(\ell)}(\cdot)$ are implemented as lightweight two-layer MLPs. The transform network is defined as
\begin{equation}
  T^{(\ell)}(\mathbf{z}) =
    \mathbf{W}^{(\ell)}_{T,2}\,
    \operatorname{Dropout}\!\left(
      \operatorname{SiLU}\!\left(
        \mathbf{W}^{(\ell)}_{T,1}\,\mathbf{z}
      \right)
    \right),
  \quad
  \mathbf{W}^{(\ell)}_{T,1} \in \mathbb{R}^{d \times h_g},\;
  \mathbf{W}^{(\ell)}_{T,2} \in \mathbb{R}^{h_g \times d},
\end{equation}
and the gate network is 
\begin{equation}
  G^{(\ell)}(\mathbf{z}) =
    \mathbf{W}^{(\ell)}_{G,2}\,
    \operatorname{SiLU}\!\left(
      \mathbf{W}^{(\ell)}_{G,1}\,\mathbf{z}
    \right),
  \quad
  \mathbf{W}^{(\ell)}_{G,1} \in \mathbb{R}^{d \times h_g},\;
  \mathbf{W}^{(\ell)}_{G,2} \in \mathbb{R}^{h_g \times d},
\end{equation}
where $h_g$ is the gate hidden size hyperparameter.
All projection matrices in both $T^{(\ell)}$ and $G^{(\ell)}$ are bias-free.

\section{Auxiliary Shadow Loss as a Regularizer}
\label{apd:auxiliary_shadow_loss}
The auxiliary shadow loss in \textcolor{blue}{Eqs.~\ref{eq:loss_lm}} and \textcolor{blue}{\ref{eq:loss_cls}} directly supervises the shadow model. Since the base model is frozen, the main loss reaches the shadow through its effects on backbone representations; the auxiliary objective provides an additional task-level learning signal to the shadow output. We treat its regularization and optimization effects as hypotheses pending a controlled ablation. Direct supervision may also affect detached shadow-only performance, which we report as a secondary analysis.

\section{Dataset Prompt Templates}
\label{apd:prompt_template}

We describe the prompt templates used to format each dataset for supervised fine-tuning.
All prompts are wrapped in the model's chat template via \texttt{apply\_chat\_template}
(with \texttt{add\_generation\_prompt=True} and \texttt{enable\_thinking=False}),
and the assistant response is appended directly after the generated prompt prefix.
Prompt tokens are masked with $-100$ in the label sequence so that loss is computed only over the answer tokens.

\paragraph{MMLU.}
Each MMLU example is formatted as a four-way multiple-choice question. The model is instructed to respond with a single letter. It supports two templates:

1) \textbf{Zero-shot template}: it is applied in the main experiment.

\begin{promptbox}
\begin{verbatim}
Question: <question>

Options:
A: <choice_0>
B: <choice_1>
C: <choice_2>
D: <choice_3>

Instructions: Answer with ONLY the letter (A, B, C, or D).
Do not include any explanation, reasoning, or additional text.
Answer:
\end{verbatim}
\end{promptbox}

\noindent Target (gold label): a single letter, e.g.\ \texttt{B}.

2) \textbf{Few-shot template}: it is used in generalization testing. When few-shot prompting is enabled, the following prefix is prepended to the zero-shot template above:

\begin{promptbox}
\begin{verbatim}
Examples of correct response format:

Question: What is 2+2?
Options:
A: 3
B: 4
C: 5
D: 6
Answer: B

Question: What color is the sky?
Options:
A: Green
B: Blue
C: Red
D: Yellow
Answer: B

Now answer the following question:

\end{verbatim}
\end{promptbox}

\paragraph{GSM8K.}
GSM8K examples are formatted for arithmetic reasoning.

\begin{promptbox}
\begin{verbatim}
Question: <question>
Answer:
\end{verbatim}
\end{promptbox}
\noindent Target: the full gold solution string, including intermediate reasoning steps and the final answer line \texttt{\#\#\#\# <number>}.

\paragraph{SQuAD~v2.}
SQuAD~v2 examples are formatted as extractive reading-comprehension tasks.
Unanswerable questions are supervised with the literal token \texttt{unanswerable}.

\begin{promptbox}
\begin{verbatim}
Answer with the exact span from the context.
If the question is unanswerable from the context,
respond with exactly: unanswerable

Context:
<context>

Question:
<question>

Answer (span or `unanswerable' only):
\end{verbatim}
\end{promptbox}
\noindent Target: the first gold answer span, or \texttt{unanswerable}.


\begin{table*}[!htbp]
  \centering
  \scriptsize
  \setlength{\tabcolsep}{4pt}
  \renewcommand{\arraystretch}{1.18}
  \caption{Case study on robot dog intent generation. Qwen3 8B serves as the base model. \textcolor{red}{Red} color  marks wrong answers. \textcolor{orange}{Orange} color  annotates hallucinations.}
  \begin{tabular}{@{}
    >{\centering\arraybackslash}p{0.04\linewidth}
    >{\raggedright\arraybackslash}p{0.12\linewidth}
    >{\raggedright\arraybackslash}p{0.18\linewidth}
    >{\raggedright\arraybackslash}p{0.18\linewidth}
    >{\raggedright\arraybackslash}p{0.18\linewidth}
    >{\raggedright\arraybackslash}p{0.18\linewidth}
  @{}}
    \toprule
    \rowcolor{black!5}
    \textbf{ID} & \textbf{User Input} & \textbf{Qwen3-0.6B Full SFT} & \textbf{Detached Shadow-Only} & \textbf{ShadowPEFT (attached shadow)} & \textbf{Qwen3-8B LoRA} \\
    \midrule
    1 & Hey dog, do a full spin, please
    & \texttt{[TurnAround()]} Full spin — here I go!
      & \texttt{[TurnAround()]} Complete rotation -- done!
      & \texttt{[TurnAround()]} Complete rotation -- done!
      & \texttt{[TurnAround()]} Full spin -- here I go!
      \\
    \midrule

    2 & Don't spleep, stand up
      & \texttt{[StandUp()]} Standing up — \textcolor{red}{I'm seated!}
      & \texttt{[StandUp()]} Up I go -- standing tall!
      & \texttt{[StandUp()]} Standing up!
      & \texttt{[StandUp()]} Good idea! Standing up right now! \\
    \midrule

    3 & How is the weather today
    & \textcolor{orange}{\texttt{[CheckWeather()]}} Checking the weather for you!
      & \texttt{\lzx{[REMOTE]}} Weather needs the internet -- routing to cloud!
      & I don't have real-time weather data! Check a weather app or website for today's forecast.
      & \textcolor{orange}{\texttt{[CheckWeather()]}} Checking the weather for you! \\
    \midrule

    4 & Hey, 19 + 18 =?
          & 19 + 18 = 37!
      & \texttt{\lzx{[REMOTE]}} Arithmetic routing to cloud!
      & 19 + 18 = 37!
      & 19 + 18 = 37! \\
    \midrule

    5 & Hey dog, tell a joke
     & \textcolor{orange}{\texttt{[RiseElevate()]}} Dog time — here I go!
      & \texttt{\lzx{[REMOTE]}} Humor needs cloud intelligence!
      & Why don't skeletons fight each other? Because they have no guts!
      & Why don't scientists trust atoms? Because they make up everything! \\

\midrule

    6 & Hey dog, play kungfu
      & \texttt{[PlayKungFu()]} Kung fu mode! Hi-ya!
      & \texttt{[PlayKungFu()]} Kung fu activated -- hi-ya!
      & \texttt{[PlayKungFu()]} Kung fu activated -- hi-ya!
      & \texttt{[PlayKungFu()]} Kung fu activated -- hi-ya! \\
    \midrule

    7 & Hey dog, please look right
      & \texttt{[TurnRight()]} Going right!
      & \texttt{[TurnRight()]} Pivoting right!
      & \texttt{[TurnRight()]} Turning right -- rotating!
      & \texttt{[TurnRight()]} Turning right -- rotating! \\
    \midrule

    8 & Hey dog, pounce forward
      & \texttt{[FrontPounce()]} Front pounce! Leaping forward!
      & \texttt{[FrontPounce()]} Pouncing -- watch out!
      & \texttt{[FrontPounce()]} Pounce activated -- here I come!
      & \texttt{[FrontPounce()]} Pounce activated -- here I come! \\
    \midrule

    9 & Explain how airplanes can fly
      & \texttt{\lzx{[REMOTE]}} Physics explanation -- cloud!
      & Airplanes fly by creating lift with their wings. As air flows faster over the curved top surface than the flat bottom, it creates a pressure difference that lifts the plane. Engines provide thrust to move forward!
      & \textcolor{orange}{\texttt{[DrawPlane()]}} Cool, drawing a plane!
      & Wings are shaped so that air moves faster over the top surface than the bottom. This creates lower air pressure above the wing and higher pressure below, which lifts the plane! \\
    \midrule

    10 & Recommend the best restaurant near me
      & \texttt{\lzx{[REMOTE]}} Restaurant recommendations need the cloud!
      & I don't have access to your location or real-time data. Try Google Maps, Yelp, or your phone's restaurant app for nearby options!
      & Sure! I'd love to recommend a great restaurant, but I need to know where you are located!
      & I don't have access to your location, but you can check Google Maps or Yelp for the best-rated restaurants near you! Just search for ``best restaurants near me'' for great suggestions. \\
    \bottomrule
  \end{tabular}
  \label{tab:case-study}
\end{table*}

\begin{table}[htbp]
  \centering
  \caption{Summary statistics of the robot-dog instruction dataset.}
  \begin{tabular}{lrrr}
    \toprule
    \textbf{Split} & \textbf{Size} & \textbf{EN} & \textbf{ZH} \\
    \midrule
    \texttt{train\_normal}        & 3400 & 2074 & 1326 \\
    \texttt{train\_remote\_base}  &  999 &  540 &  459 \\
    \texttt{train\_remote\_shadow}&   47 &   21 &   26 \\
    \texttt{test\_normal}         &  306 &  170 &  136 \\
    \texttt{test\_remote}         &   18 &   10 &    8 \\
    \midrule
    \textbf{Total}                & 4770 & 2815 & 1955 \\
    \bottomrule
  \end{tabular}
  \label{tab:robot-dataset-summary}
\end{table}

\section{High-Capacity Shadow Initialization}
\label{apd:shadow_pretraining}
This section describes the separate high-capacity experiment in which a roughly 0.5B parameter shadow is attached to Qwen3 8B. This configuration has 454,991,872 trainable parameters and is not comparable in trainable-parameter budget to the compact ShadowPEFT, LoRA, and DoRA configurations in the primary experiments.

Due to resource constraints, we reuse Qwen3 0.6B as the shadow model $f_\theta$ rather than pretraining a new model from scratch. To attach it to Qwen3 8B, we bridge their hidden dimensions $d_s \neq d_t$ via a linear projection $\mathbf{P} \in \mathbb{R}^{d_t \times d_s}$, giving the forward pass:
\begin{equation}
    \hat{\mathbf{y}} = \mathbf{W}_{\mathrm{lm}}\,\mathbf{P}\,\mathbf{h},
    \label{eq:forward}
\end{equation}
\noindent where $\mathbf{h} \in \mathbb{R}^{d_s}$ are the shadow hidden states and $\mathbf{W}_{\mathrm{lm}} \in \mathbb{R}^{V \times d_t}$ is the frozen 8B LM head. A randomly initialised $\mathbf{P}$ destroys the model's generation ability. To recover a useful starting point, we initialise $\mathbf{P}$ by minimising the Frobenius distance to Qwen3 0.6B's original head $\mathbf{W}_{\mathrm{lm}}^{\mathrm{ref}}$:
\begin{equation}
    \mathbf{P}^* = \mathbf{W}_{\mathrm{lm}}^{+}\,\mathbf{W}_{\mathrm{lm}}^{\mathrm{ref}},
    \label{eq:pinv}
\end{equation}
\noindent where $\mathbf{W}_{\mathrm{lm}}^{+}$ is the Moore--Penrose pseudo-inverse \citep{barata2012moore}. This warm start ensures the composed head $\mathbf{W}_{\mathrm{lm}}\mathbf{P}^*$ approximates the original output distribution from the first step, reducing subsequent alignment training. We then continue pretraining $f_\theta$ and $\mathbf{P}$ on FineWeb-Edu (English subset, sampled $100K$) \citep{lozhkov2024fineweb-edu} and Wudao (Chinese only, sampled $100K$) \citep{wudao} corpus with causal language modeling objective.

\section{System-Level Edge-Cloud Collaboration Evaluation}
\label{apd:edge_cloud_study}
In the main content, we compared ShadowPEFT with standard PEFT baselines under conventional settings. As discussed, ShadowPEFT's shadow module can be detached, making it particularly suitable for edge–cloud collaboration. 
To evaluate this scenario, we conduct a system-level experiment on the Unitree Go2 robot dog intent-understanding task. This task involves classifying and routing commands from a natural-language interface. 
We compare ShadowPEFT (455M trainable parameters) with two widely adopted PEFT baselines, LoRA and DoRA (each with 31M trainable parameters), and also with a small fully fine-tuned model (Qwen3-0.6B SFT) to better illustrate ShadowPEFT's detachable advantage. Note that scaling up the trainable parameter budget for LoRA or DoRA leads to degraded performance (discussed in Section \ref{sec:scaling}); therefore, we use the same small parameter count as in \textcolor{blue}{Table~\ref{tab:main_exp}}.

\paragraph{Dataset and Split.}
We release a bilingual corpus containing $4,770$ examples covering $34$ predefined robot skills and one \texttt{[REMOTE]} category. The predefined robot skills are:
\texttt{Damp()}, \texttt{BalanceStand()}, \texttt{StopMove()},
\texttt{StandUp()}, \texttt{StandDown()}, \texttt{RecoveryStand()},
\texttt{Sit()}, \texttt{RiseSit()}, \texttt{Stretch()}, \texttt{Wallow()},
\texttt{Scrape()}, \texttt{FrontFlip()}, \texttt{FrontJump()},
\texttt{FrontPounce()}, \texttt{WiggleHips()}, \texttt{TurnLeft()},
\texttt{TurnRight()}, \texttt{SayHello()}, \texttt{Dance()},
\texttt{DrawHeart()}, \texttt{BalanceAttitude()}, \texttt{PlayKungFu()},
\texttt{HappyBirthday()}, \texttt{HappyNewYear()}, \texttt{CheerLeading()},
\texttt{ShakeBody()}, \texttt{TurnAround()}, \texttt{LionDance()},
\texttt{Welcome()}, \texttt{WaltzDance()}, \texttt{ChaChaDance()},
\texttt{Tango()}, \texttt{HipHopDance()}, and \texttt{Bark()}.
The corpus was initially generated with Gemini and manually refined by one Ph.D.\ student and one research assistant. A normalized-content leakage audit revealed one exact prompt duplicate (\texttt{teleport}) appearing in both \texttt{train\_remote\_base} and \texttt{test\_remote}; we removed the training copy before any run. The corrected corpus therefore contains $4,770$ examples, as detailed in \textcolor{blue}{Table~\ref{tab:robot-dataset-summary}}. The held-out evaluation set remains $324$ examples: $306$ routine-skill examples and $18$ remote examples. No held-out example is used for training, threshold selection, accuracy, or latency evaluation.

\paragraph{ShadowPEFT Routing Protocol.}
The detached shadow first predicts either one of the predefined local skills or \texttt{[REMOTE]}. A local prediction is executed by the edge policy; \texttt{[REMOTE]} forwards the request to the attached cloud model. 

\paragraph{Results.}
\textcolor{blue}{Figure~\ref{fig:downstream}} shows the accuracy versus inference time trade-off. Directly full-parameter fine-tuning Qwen3~0.6B achieves the lowest latency but lower accuracy than Qwen3~8B fine-tuned with PEFT. Among the PEFT methods, ShadowPEFT attains both the lowest latency ($1.13\times$ and $9.98\times$ faster than LoRA and DoRA, respectively) and the highest accuracy. The low latency stems from its detachable design: routine robot skills are resolved locally by the detached shadow without accessing the cloud. The cloud model (full ShadowPEFT) is only invoked for complex or open-domain requests. This design improves real-time interaction and reduces cloud usage. Moreover, compared to the full-finetuned Qwen3~0.6B model, ShadowPEFT trains a single model that simultaneously produces both a lightweight edge module (the detached shadow) and a powerful cloud model (the full ShadowPEFT). This co-training strategy enables greater flexibility: a single training run delivers two differently capable models tailored for edge and cloud scenarios, respectively.

\paragraph{Case Studies.}
\textcolor{blue}{Table~\ref{tab:case-study}} presents several qualitative examples. The detached shadow model reliably understands routine robot skills. For more complex or open-domain queries, instead of hallucinating, it returns the \texttt{[REMOTE]} tag, correctly forwarding the input to the full ShadowPEFT model deployed on the cloud server. This behavior confirms that the detached mode serves as an effective lightweight front-end for simple instructions while conservatively escalating harder cases to the cloud, benefiting edge-computing scenarios.
The full ShadowPEFT model performs well on both routine skills and complex user queries. In contrast, although LoRA and DoRA generate fluent responses, they occasionally produce incorrect actions or hallucinations. These observations further validate the effectiveness of ShadowPEFT and the practicality of its detachable and attached modes.
Overall, ShadowPEFT provides more accurate and precise responses than LoRA and DoRA while maintaining a scalable design suitable for real-world edge–cloud deployment.

\begin{figure}
    \centering
    \begin{tikzpicture}[
  line cap=round,
  line join=round,
  >=stealth,
  font=\fontfamily{ptm}\selectfont,
  lbl/.style={font=\footnotesize, text=labelgray, align=center},
]

\pgfmathsetmacro{\xmin}{0}
\pgfmathsetmacro{\ymin}{96.5}
\pgfmathsetmacro{\ymax}{100.0}
\pgfmathsetmacro{\yspan}{\ymax-\ymin}

\pgfmathsetmacro{\plotW}{7.35}
\pgfmathsetmacro{\plotH}{5.25}
\pgfmathsetmacro{\ox}{1.38}
\pgfmathsetmacro{\oy}{1.08}

\pgfmathsetmacro{\segA}{0.16*\plotW} 
\pgfmathsetmacro{\segB}{0.24*\plotW} 
\pgfmathsetmacro{\segC}{0.28*\plotW} 
\pgfmathdeclarefunction{xmap}{1}{%
  \pgfmathparse{#1 <= 100 ? (#1/100)*\segA :
    (#1 <= 200 ? \segA + ((#1-100)/100)*\segA :
    (#1 <= 300 ? 2*\segA + ((#1-200)/100)*\segA :
    (#1 <= 1000 ? 3*\segA + ((#1-300)/700)*\segB :
    3*\segA + \segB + ((#1-1000)/1000)*\segC)))}
}

\pgfmathsetmacro{\pxS}{xmap(116.756)}
\pgfmathsetmacro{\pyS}{(99.35-\ymin)/\yspan*\plotH}
\pgfmathsetmacro{\pxB}{xmap(248.306)}
\pgfmathsetmacro{\pyB}{(97.7-\ymin)/\yspan*\plotH}
\pgfmathsetmacro{\pxC}{xmap(1281.669)}
\pgfmathsetmacro{\pyC}{(97.7-\ymin)/\yspan*\plotH}
\pgfmathsetmacro{\pxQ}{xmap(101)}
\pgfmathsetmacro{\pyQ}{(97.4-\ymin)/\yspan*\plotH}

\pgfmathsetmacro{\sx}{\ox+\pxS}
\pgfmathsetmacro{\sy}{\oy+\pyS}
\pgfmathsetmacro{\bx}{\ox+\pxB}
\pgfmathsetmacro{\by}{\oy+\pyB}
\pgfmathsetmacro{\cx}{\ox+\pxC}
\pgfmathsetmacro{\cy}{\oy+\pyC}
\pgfmathsetmacro{\qx}{\ox+\pxQ}
\pgfmathsetmacro{\qy}{\oy+\pyQ}

\pgfmathsetmacro{\canL}{-0.15}
\pgfmathsetmacro{\canR}{\ox+\plotW+1.05}
\pgfmathsetmacro{\canB}{-0.75}
\pgfmathsetmacro{\canT}{\oy+\plotH+1.12}
\fill[white] (\canL,\canB) rectangle (\canR,\canT);

\foreach \t in {0,100,200,300,1000,1500,2000} {
  \pgfmathsetmacro{\gx}{xmap(\t)}
  \draw[color=gridline, very thin] ({\ox+\gx},\oy) -- ({\ox+\gx},{\oy+\plotH});
}
\foreach \a in {96.5,97.0,97.5,98.0,98.5,99.0,99.5,100.0} {
  \pgfmathsetmacro{\gy}{(\a-\ymin)/\yspan*\plotH}
  \draw[color=gridline, very thin] (\ox,{\oy+\gy}) -- ({\ox+\plotW},{\oy+\gy});
}

\draw[color=axiscolor, semithick] (\ox,\oy) rectangle ({\ox+\plotW},{\oy+\plotH});

\foreach \t in {0,100,200,300,1000,1500,2000} {
  \pgfmathsetmacro{\gx}{xmap(\t)}
  \draw[color=axiscolor] ({\ox+\gx},\oy) -- ({\ox+\gx},{\oy-0.06});
  \node[font=\scriptsize, text=subgray, below=2pt] at ({\ox+\gx},{\oy-0.06}) {\t};
}
\node[font=\small\bfseries, text=labelgray, below=15pt] at ({\ox+\plotW/2},\oy)
  {Inference time (s)};

\foreach \a in {96.5,97.0,97.5,98.0,98.5,99.0,99.5,100.0} {
  \pgfmathsetmacro{\gy}{(\a-\ymin)/\yspan*\plotH}
  \draw[color=axiscolor] ({\ox-0.06},{\oy+\gy}) -- (\ox,{\oy+\gy});
  \node[font=\scriptsize, text=subgray, anchor=east, xshift=-3pt] at ({\ox-0.06},{\oy+\gy})
    {\pgfmathprintnumber[fixed,precision=1]{\a}};
}
\node[font=\small\bfseries, text=labelgray, rotate=90, anchor=south] at ({\ox-0.92},{\oy+\plotH/2})
  {Accuracy (\%)};

\def\smallbubble{0.35}
\def\middlebubble{0.45}
\def\bigbubble{0.55}
\def\tinybubble{0.3}
\def\bubbleline{0.18pt}

\filldraw[fill=bubbleA, draw=bubbleA!55!black, line width=\bubbleline]
  (\sx,\sy) circle (\bigbubble);
\node[text=black, font=\scriptsize\bfseries] at (\sx,\sy) {\pgfmathprintnumber[fixed,precision=2]{99.35}};
\node[lbl, anchor=north, text width=4.0cm, inner sep=1pt] at ($(\sx,\sy)+(0.0,0.9)$)
  {\textbf{Qwen3 8B (ShadowPEFT)}};

\filldraw[fill=bubbleB, draw=bubbleB!55!black, line width=\bubbleline]
  (\bx,\by) circle (\middlebubble);
\node[text=black, font=\scriptsize\bfseries] at (\bx,\by) {\pgfmathprintnumber[fixed,precision=1]{97.7}};
\node[lbl, anchor=north, text width=3cm, inner sep=1pt] at ($(\bx,\by)+(0,+0.8)$)
  {Qwen3 8B (LoRA)};

\filldraw[fill=bubbleB, draw=bubbleC!55!black, line width=\bubbleline]
  (\cx,\cy) circle (\smallbubble);
\node[text=black, font=\scriptsize\bfseries] at (\cx,\cy) {\pgfmathprintnumber[fixed,precision=1]{97.7}};
\node[lbl, anchor=north, text width=3cm, inner sep=1pt] at ($(\cx,\cy)+(0,0.8)$)
  {Qwen3 8B (DoRA)};

\filldraw[fill=bubbleB, draw=bubbleB!55!black, line width=\bubbleline]
  (\qx,\qy) circle (\tinybubble);
\node[text=black, font=\scriptsize\bfseries] at (\qx,\qy) {\pgfmathprintnumber[fixed,precision=1]{97.4}};
\node[lbl, anchor=north, text width=3.0cm, inner sep=1pt] at ($(\qx,\qy)+(0,0.7)$)
  {Qwen3 0.6B SFT};
\end{tikzpicture}
    \vspace{-25px}
    \caption{Accuracy vs.\ inference time on a test set with $4{,}770$ samples.}
    \label{fig:downstream}
\end{figure}

\section{Experimental Setup for Subject-Driven Image Generation}
\label{apd:dino}
We evaluate ShadowPEFT on the DreamBooth benchmark
\citep{ruiz2023dreambooth}, which contains 30 subjects, each paired with 25 evaluation prompts. Following \citet{ruiz2023dreambooth} and \citet{bini2025decoupling}, we fine-tune Stable Diffusion \citep{rombach2022high} to generate each reference subject in the context specified by a given prompt. For every subject--prompt pair, we generate four images after fine-tuning and evaluate subject fidelity using DINO \citep{caron2021emerging} and CLIP-I \citep{radford2021learning}. Both metrics measure similarity between generated and reference images, with higher scores indicating stronger preservation of the target subject. DINO is treated as the primary subject-fidelity metric because it is particularly sensitive to subject-specific visual features \citep{ruiz2023dreambooth}. We compare ShadowPEFT against the widely used LoRA \citep{hu2022lora} and DoRA \citep{liu2024dora}, as well as DeLoRA \citep{bini2025decoupling}, the previous state of the art for this task. Following \citet{bini2025decoupling}, tuned hyperparameters are used for all PEFT methods. The reported ShadowPEFT scores are averaged over five independent runs.

\end{document}